\documentclass[11pt, letterpaper]{article}
\usepackage{arxiv}

\RequirePackage[hyphens]{url}
\usepackage[export]{adjustbox}
\usepackage[ruled,vlined, linesnumbered]{algorithm2e}
\usepackage{algpseudocode}
\usepackage{alltt}
\usepackage{amsmath}
\usepackage{amssymb}
\usepackage{animate}
\usepackage{array}
\usepackage{authblk}
\usepackage{bm}
\usepackage{booktabs}
\usepackage{caption}
\usepackage{capt-of}
\usepackage{changepage}
\usepackage{color}
\usepackage{csvsimple}
\usepackage{enumitem}
\usepackage{etoolbox}
\usepackage{fontawesome5}
\usepackage[T1]{fontenc}
\usepackage{framed}

\usepackage{graphicx}
\usepackage[colorlinks=true, linkcolor = blue, citecolor = blue, urlcolor = blue]{hyperref}
\usepackage[utf8]{inputenc}

\usepackage{longtable}
\usepackage{lscape}
\usepackage{makecell}
\usepackage{multicol}
\usepackage{multirow}
\usepackage[numbers, round, sort&compress]{natbib}
\usepackage{realboxes}
\usepackage{rotating}
\usepackage{setspace}
\usepackage{siunitx}

\usepackage{soul}
\usepackage{subcaption}
\usepackage{svg}
\usepackage[most]{tcolorbox}
\usepackage{todonotes}
\tcbuselibrary{listings}

\usepackage{tikz}
\usetikzlibrary{shapes, arrows.meta, positioning, decorations.pathreplacing}

\usepackage{url}
\usepackage{xcolor}

\graphicspath{ {Figures/} }
\newcolumntype{C}[1]{>{\centering\arraybackslash}m{#1}}
\newcommand{\instructions}[1]{}

{ \BeforeBeginEnvironment{tabular}{\begin{center}\footnotesize}
\AfterEndEnvironment{tabular} {\end{center}} }

\DeclareMathOperator*{\argmax}{arg\,max}

\definecolor{light-gray}{gray}{0.96} 
\definecolor{dark-gray}{gray}{0.8}

\lstnewenvironment{VerbItem}{%
    \lstset{basicstyle=\ttfamily\small,breakatwhitespace,breaklines,
            numbers=left,numbersep=2pt,xleftmargin=0.75em}}{}

\newif\ifblinded
\newif\iffinal

\newcommand{\setfinal}{
    \blindedfalse
    \finaltrue
}

\newcommand{\authorinfo}{
    \ifblinded
        \author{\large (Authors blinded for peer review)}
    \fi
    \iffinal
       \author[1]{Fadel M. Megahed}
\author[2]{Ying-Ju Chen}
\author[3]{Bianca Maria Colosimo}
\author[3]{Marco Luigi Giuseppe Grasso}
\author[1]{L. Allison Jones-Farmer}
\author[4]{Sven Knoth}
\author[5]{Hongyue Sun}
\author[6,*]{Inez Zwetsloot}

\affil[1]{Farmer School of Business, Miami University, Oxford OH, USA.}
\affil[2]{College of Arts and Sciences, University of Dayton, Dayton OH, USA.}
\affil[3]{Department of Mechanical Engineering, Politecnico di Milano, Milan, Italy.}
\affil[4]{Mathematics \& Statistics, Helmut Schmidt University, Hamburg, Germany.}
\affil[5]{College of Engineering, University of Georgia, Athens, GA, USA.}
\affil[6]{Amsterdam Business School, University of Amsterdam, Amsterdam, The Netherlands.}
\affil[*]{Corresponding author. Email: \href{mailto:I.M.Zwetsloot@uva.nl}{I.M.Zwetsloot@uva.nl}}

    \fi
}

\usepackage[utf8]{inputenc}

\usepackage{authblk}
\usepackage{ulem}
\usepackage{listings} 
\usepackage{comment}

\title{\Large{\textbf{Adapting OpenAI's CLIP Model for Few-Shot Image Inspection in Manufacturing Quality Control: An Expository Case Study with Multiple Application Examples}}}

\setfinal

\date{\small \today}

\renewcommand{\shorttitle}{Few-Shot Image Inspection with CLIP}

\begin{document}

\authorinfo 

\maketitle 

\begin{abstract}
\vspace{\baselineskip}
\noindent This expository paper introduces a simplified approach to image-based quality inspection in manufacturing using OpenAI's CLIP (Contrastive Language-Image Pretraining) model adapted for few-shot learning. While CLIP has demonstrated impressive capabilities in general computer vision tasks, its direct application to manufacturing inspection presents challenges due to the domain gap between its training data and industrial applications. We evaluate CLIP's effectiveness through five case studies: metallic pan surface inspection, 3D printing extrusion profile analysis, stochastic textured surface evaluation, automotive assembly inspection, and microstructure image classification. Our results show that CLIP can achieve high classification accuracy with relatively small learning sets (50-100 examples per class) for single-component and texture-based applications. However, the performance degrades with complex multi-component scenes. We provide a practical implementation framework that enables quality engineers to quickly assess CLIP's suitability for their specific applications before pursuing more complex solutions. This work establishes CLIP-based few-shot learning as an effective baseline approach that balances implementation simplicity with robust performance, demonstrated in several manufacturing quality control applications.  
\end{abstract}

\vspace{\baselineskip}



\doublespacing


\section{Introduction}
\label{sec:intro}
The monitoring and inspection of image data in manufacturing environments presents a fundamental dichotomy in statistical quality monitoring. Traditional approaches, primarily leveraging control charts, have provided rigorous theoretical frameworks for detecting process shifts. Concurrently, computational approaches ranging from traditional image processing to modern deep learning architectures have demonstrated remarkable empirical success. This dichotomy has led to two distinct schools of thought, each with inherent strengths and limitations.

The statistical process monitoring school, grounded in control chart theory, offers two compelling advantages. First, these methods require only in-control data for implementation, circumventing the often insurmountable challenge of collecting a representative sample of defective products. Second, they provide theoretical guarantees through their ability to aggregate information temporally, offering insights into shift location, timing, and magnitude. However, the practical implementation of such methods faces significant challenges. They typically require extensive in-control samples, verification of process stability, and specific mathematical formulations closely tied to particular quality data model assumptions, often resulting in methods tailored to specific applications with uncertain generalizability. Furthermore, the lack of accessible software packages poses a general challenge.
When applied to image and video image data, traditional approaches typically begin with image preprocessing to extract specific features, followed by control charting to assess process stability over time. Stability can also be evaluated using functional or profile data, representing the in-control dynamics of the video image within the region of interest \citep{megahed2011review,  menafoglio2018profile, colosimo2018modeling, colosimo2018opportunities}. Spatio-temporal modeling has proven effective for identifying out-of-control states, although it introduces additional complexity in the mathematical formulation \citep{yan2018real}.

The computational school, conversely, has evolved from basic image-processing techniques to sophisticated deep-learning architectures. Early works, such as \citet{megahed2012real}, demonstrated the utility of relatively simple image processing approaches. The field then witnessed a dramatic shift with the success of deep learning architectures, beginning with \citet{krizhevsky2012imagenet}'s \texttt{AlexNet}, followed by increasingly sophisticated networks such as \texttt{VGG} \citep{simonyan2015very}, \texttt{ResNet} \citep{he2016deep}, and \texttt{EfficientNet} \citep{tan2019efficientnet}. While powerful, these modern deep-learning methods often require substantial datasets for fine-tuning and optimization. Thus, such methods demand significant expertise in machine learning and domain-specific knowledge. In our estimation, the statistical community's skepticism toward these approaches stems from their empirical nature, extensive data requirements (for both in-control and out-of-control cases, which the only exception of methods based on the one-class classifier), and inability to accumulate evidence over time—though this limitation is primarily relevant when detection sensitivity is suboptimal. Note that the need for large datasets remains true for applications which combine statistical control charts with deep learning methods; for example, \citet{kang2024modern} used an augmented training set of 16,850 in-control and 7,410 out-of-control images. 

\citet{okhrin2024control} recently noted that most image process monitoring procedures rely on aggregated image characteristics, such as entropy or arithmetic averages, due to pixel-level analysis's computational and theoretical challenges. This observation parallels the fundamental design of vision transformer models, which inherently operate on aggregated patch-level features rather than individual pixels. This architectural similarity suggests that vision transformer models might naturally align with image quality control applications while addressing some limitations across inspection and monitoring applications.

The emergence of OpenAI's \texttt{CLIP} (Contrastive Language-Image Pretraining) model \citep{radford2021learning} presents an intriguing opportunity to address these challenges. Trained on 400 million image-text pairs, \texttt{CLIP}'s dual-encoder architecture has demonstrated exceptional feature extraction capabilities across various domains. However, our empirical investigations reveal that while \texttt{CLIP} often fails as a zero-shot classifier in manufacturing applications (due to the domain gap between its training data and industrial applications), its performance as a few-shot classifier is remarkably robust. This finding suggests that \texttt{CLIP} could serve as an initial benchmark for image inspection and monitoring applications, potentially offering:
\begin{itemize}[nosep]
    \item Reduced data requirements compared to traditional deep learning approaches
    \item More generalizable performance across different manufacturing contexts
    \item Natural handling of high-dimensional image data through its transformer architecture
\end{itemize}
These characteristics make \texttt{CLIP} appealing for quality engineering practitioners seeking robust, yet implementable, solutions for image-based inspection tasks.

This paper systematically evaluates \texttt{CLIP}'s utility in manufacturing quality control through few-shot learning. Following \citet{megahed2024comparing}'s argument that proper benchmarking is essential before deploying complex methods, we position \texttt{CLIP} as a simple yet powerful baseline that should be evaluated before implementing more sophisticated approaches. Our proposition is straightforward: if \texttt{CLIP}'s few-shot learning capabilities prove sufficient for a given application, there may be no need for more complex methods, especially since our approach can be implemented with minimal code and computational overhead. Specifically, the objectives of this study are threefold:
\begin{enumerate}[label=(\arabic*), nosep]
    \item To demonstrate how \texttt{CLIP} can be effectively adapted using few-shot learning for manufacturing quality control. We provide a practical framework that reduces implementation complexity while maintaining high accuracy. We emphasize that this framework is a baseline that should be tested before pursuing more complex solutions.
    \item To investigate the relationship between learning set size and classification performance, and offer guidance on the minimal data requirements for effective implementation.
    \item To examine the impact of vision transformer model selection (by comparing \texttt{ViT-L/14} and \texttt{ViT-B/32}) on classification accuracy and computational efficiency.
\end{enumerate}
Through these objectives, we aim to establish a systematic framework for evaluating whether this simplified approach can meet practitioners' needs before investing in more complex solutions.

To examine the utility of our approach, we present five diverse case studies: metallic pan surface inspection \citep{megahed2012real}, 3D printing extrusion profile analysis, stochastic textured surface evaluation \citep{bui2018monitoring,bui2021spc4sts},  automotive assembly inspection \citep{carvalho2024detecting}, and microstructure image classification for metal additive manufacturing \citep{wei2025lowdimensional}. These cases represent a broad spectrum of manufacturing quality control challenges, from well-defined defect patterns to subtle variations in surface texture. Through these examples, we demonstrate that \texttt{CLIP}-based few-shot learning can offer a practical middle ground between traditional statistical process monitoring and modern computational approaches. Our examples indicate that \texttt{CLIP}-based few-shot learning can provide a new paradigm for image-based quality control in manufacturing environments due to its predictive performance and implementation simplicity.

\section{Background}
\label{sec:background}
\subsection{The \texttt{CLIP} Model Architecture}

OpenAI's \texttt{CLIP} (Contrastive Language-Image Pretraining) represents a significant advancement in multimodal learning. Pretrained on more than 400 million image-text pairs, \texttt{CLIP} employs a dual-encoder architecture that simultaneously processes visual and textual inputs. The uses of the \texttt{CLIP} model are varied and include image search and retrieval, content moderation, text-to-image and image-to-text generation, and zero-shot image classification \citep{radford2021learning}. At its core, \texttt{CLIP} consists of an image encoder $f_{\theta}$ and a text encoder $g_{\phi}$, where $\theta$ and $\phi$ represent their respective parameters. These encoders map images and text into a shared high-dimensional embedding space $\mathbb{R}^d$, where $d$ varies by encoder model. Table \ref{tab:clip_variants} captures some popular image encoder models that can be used within the \texttt{CLIP} architecture.

\begin{table}[ht]
\begin{adjustwidth}{-0.8in}{-0.8in}
\centering
\caption{Popular image encoder models used with the \texttt{CLIP} model and their characteristics.}
\label{tab:clip_variants}
\begin{tabular}{C{0.6in}p{1.6in}C{0.7in}C{0.75in}C{0.7in}p{1.1in}C{0.8in}}
\toprule
\textbf{Encoder Model} & \textbf{Architectural Details} & \textbf{Embedding Dim.} & \textbf{Compute Efficiency} & \textbf{Detail Capturing} & \textbf{Typical Use Case} & \textbf{Linked Model Card} \\
\midrule

\texttt{ViT-B/32} & Splits image into 32×32 pixel patches, trading spatial resolution for speed & 512 & High & Moderate & Simple tasks, small datasets, limited resources & \href{https://huggingface.co/openai/clip-vit-base-patch32}{ViT-B/32} \citep{openai2021clipmodelcard} \\[8ex]

\texttt{ViT-B/16} & Uses 16×16 pixel patches, offering better spatial resolution than ViT-B/32 & 512 & Moderate & High & Medium complexity tasks, more variability & \href{https://huggingface.co/openai/clip-vit-base-patch16}{ViT-B/16} \citep{openai2021clipmodelcard} \\[8ex]

\texttt{ViT-L/14} & Uses 14×14 pixel patches and larger model capacity, maximizing spatial detail & 768 & Low & Very High & Complex tasks, large datasets, ample resources & \href{https://huggingface.co/openai/clip-vit-large-patch14}{ViT-L/14} \citep{openai2021clipmodelcard} \\[8ex]

\texttt{RN50} & Standard 50-layer ResNet architecture with basic scaling & 1024 & High & Moderate & Legacy compatibility, fast inference & \href{https://github.com/openai/CLIP/blob/main/model-card.md}{RN50} \citep{openai2021clipmodelcard} \\[6ex]

\texttt{RN50x16} & 50-layer ResNet with 16x wider layers, offering increased model capacity & 768 & Moderate & High & Detailed tasks, medium resources & \href{https://github.com/openai/CLIP/blob/main/model-card.md}{RN50x16} \citep{openai2021clipmodelcard} \\[8ex]
\bottomrule
\end{tabular}
\end{adjustwidth}
\end{table}

\begin{figure}[ht]
\centering
\includegraphics[width=0.98\textwidth, frame]{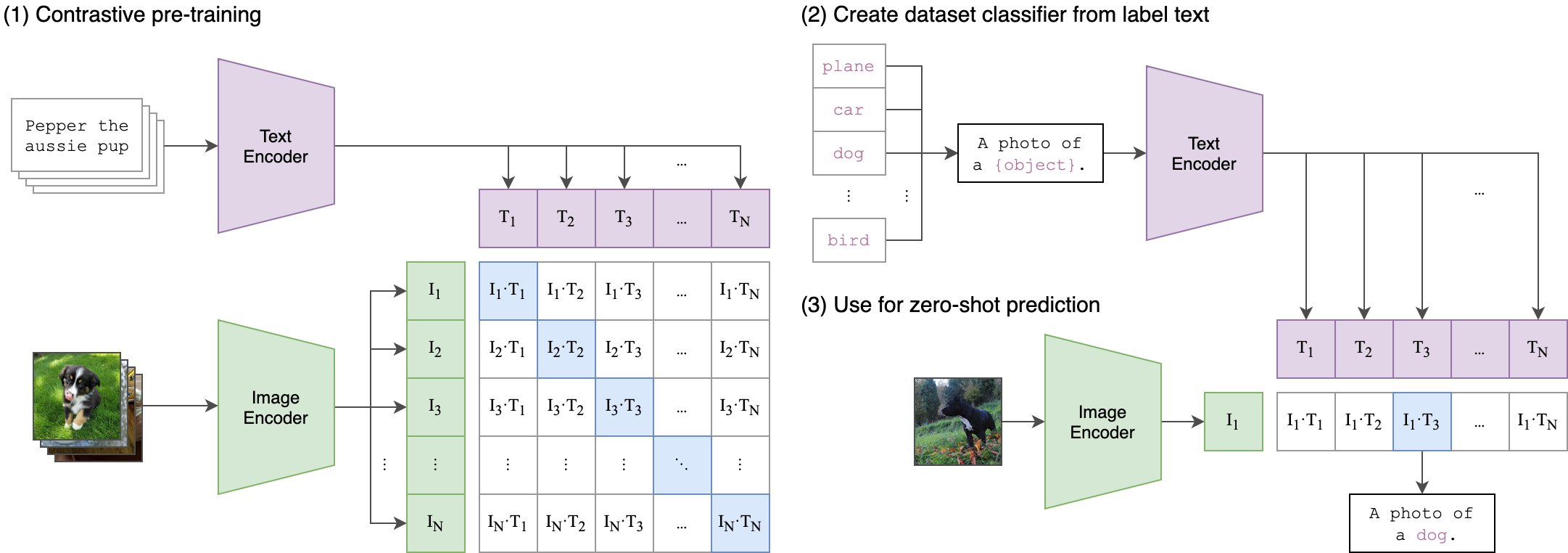}
\caption{CLIP's architecture and workflow: (1) the model learns to align image and text embeddings through contrastive learning during pretraining. (2) text templates are created from class labels for classification tasks. (3) Zero-shot prediction compares image embeddings with text embeddings of possible classes. Image source: OpenAI, provided through an MIT license at \url{https://github.com/openai/CLIP/blob/main/CLIP.png}.}
\label{fig:clip_architecture}
\end{figure}

The \texttt{CLIP} architecture, depicted in Figure \ref{fig:clip_architecture}, operates through three distinct stages. During pretraining, both image and text encoders transform millions of image-text pairs into a shared embedding space. The image encoder first standardizes inputs to model-specific dimensions; typically $224\times224$ pixels for models like \texttt{ViT-B/32} and \texttt{ViT-B/16}, or $336\times336$ pixels for \texttt{ViT-L/14} to capture finer details. Simultaneously, the text encoder processes corresponding textual descriptions and creates embeddings that enable the model to learn semantic relationships between visual and textual content. Both the image and text encoders produce embeddings of the same dimensionality, ensuring that the outputs can be directly compared in the shared embedding space. This alignment enables the model to learn semantic relationships between visual and textual content effectively.

\subsection{Mathematical Foundations}

Input images undergo critical preprocessing steps before entering the \texttt{CLIP} encoders. For an input image $x \in \mathbb{R}^{H \times W \times 3}$ of arbitrary height $H$ and width $W$, the preprocessing function $p(\cdot)$ performs:

\begin{equation}
x' = p(x) = \text{Resize}(\text{CenterCrop}(x, s), s)
\end{equation}

\noindent where $s$ is the model-specific input size. For instance, an image of size $3264\times2448$ is first center cropped to $2448\times2448$ (i.e., the smaller dimension to create a square image) before being resized to $336\times336$ for the \texttt{ViT-L/14} model. This two-step process helps preserve the proportions of the central region while standardizing the input size.

This standardized input is then divided into non-overlapping patches for Vision Transformer (ViT) models. Given patch size $P$ (e.g., $32\times32$ for \texttt{ViT-B/32} or $14\times14$ for \texttt{ViT-L/14}), the image is segmented into a sequence of $N = (s/P)^2$ patches, each flattened into a vector:
\begin{equation}
\{x'_1, x'_2, ..., x'_N\} \quad \text{where} \quad x'_i \in \mathbb{R}^{3P^2}.
\end{equation}

The image encoder $f_{\theta}$ and text encoder $g_{\phi}$ transform their inputs into a shared embedding space $\mathbb{R}^d$. These embeddings are learned representations that capture semantic meaning in dense vectors. \texttt{CLIP}'s key innovation lies in jointly training these encoders so that semantically similar concepts end up close together in the embedding space, regardless of their modality (image or text).

For comparing embeddings, \texttt{CLIP} uses cosine similarity because it (a) normalizes for vector magnitude, focusing on the directional similarity; (b) bounds similarity scores between -1 and 1, providing numerical stability; and (c) remains differentiable, enabling gradient-based optimization. For a given image embedding $f_{\theta}(x)$ and text embedding $g_{\phi}(y)$, their cosine similarity is:

\begin{equation}
s(x,y) = \text{cosSim}\big(f_{\theta}(x), g_{\phi}(y)\big) = \frac{f_{\theta}(x) \cdot g_{\phi}(y)}{\|f_{\theta}(x)\| \|g_{\phi}(y)\|}.
\end{equation}

\noindent During training, \texttt{CLIP} optimizes these similarities directly through a contrastive loss function. While the original \texttt{CLIP} implementation includes a temperature parameter $\tau$ for scaling similarities during training, the core functionality relies on the raw similarities themselves.

\texttt{CLIP} enables both zero-shot and few-shot classification approaches. In zero-shot classification, the model directly compares an image embedding with text embeddings of possible classes:

\begin{equation}
c^* = \argmax_{c \in \mathcal{C}} \text{cosSim}\big(f_{\theta}(x), g_{\phi}(t_c)\big)
\end{equation}

\noindent where $t_c$ represents the text template for class $c$, and $\mathcal{C}$ is the set of possible classes. For few-shot classification, given a set of example images $\{x_k^c\}_{k=1}^K$ for each class $c$, we compute the maximum similarity between a new image $x$ and all examples within each class:
\begin{equation}
s_c(x) = \max_{k \in {1,\ldots,K}} \text{cosSim}\big(f_{\theta}(x), f_{\theta}(x_k^c)\big).
\end{equation}
\noindent New images are then classified based on the class with the highest maximum similarity:
\begin{equation}
c^* = \argmax_{c \in \mathcal{C}} s_c(x).
\end{equation}

\noindent One can then convert these similarities into a probability using the classical \texttt{softmax} transformation \citep{bridle1990probabilistic}. 

\subsection{An Illustrative Example}

\begin{figure}[htb!]
\centering
\includegraphics[width=0.475\textwidth]{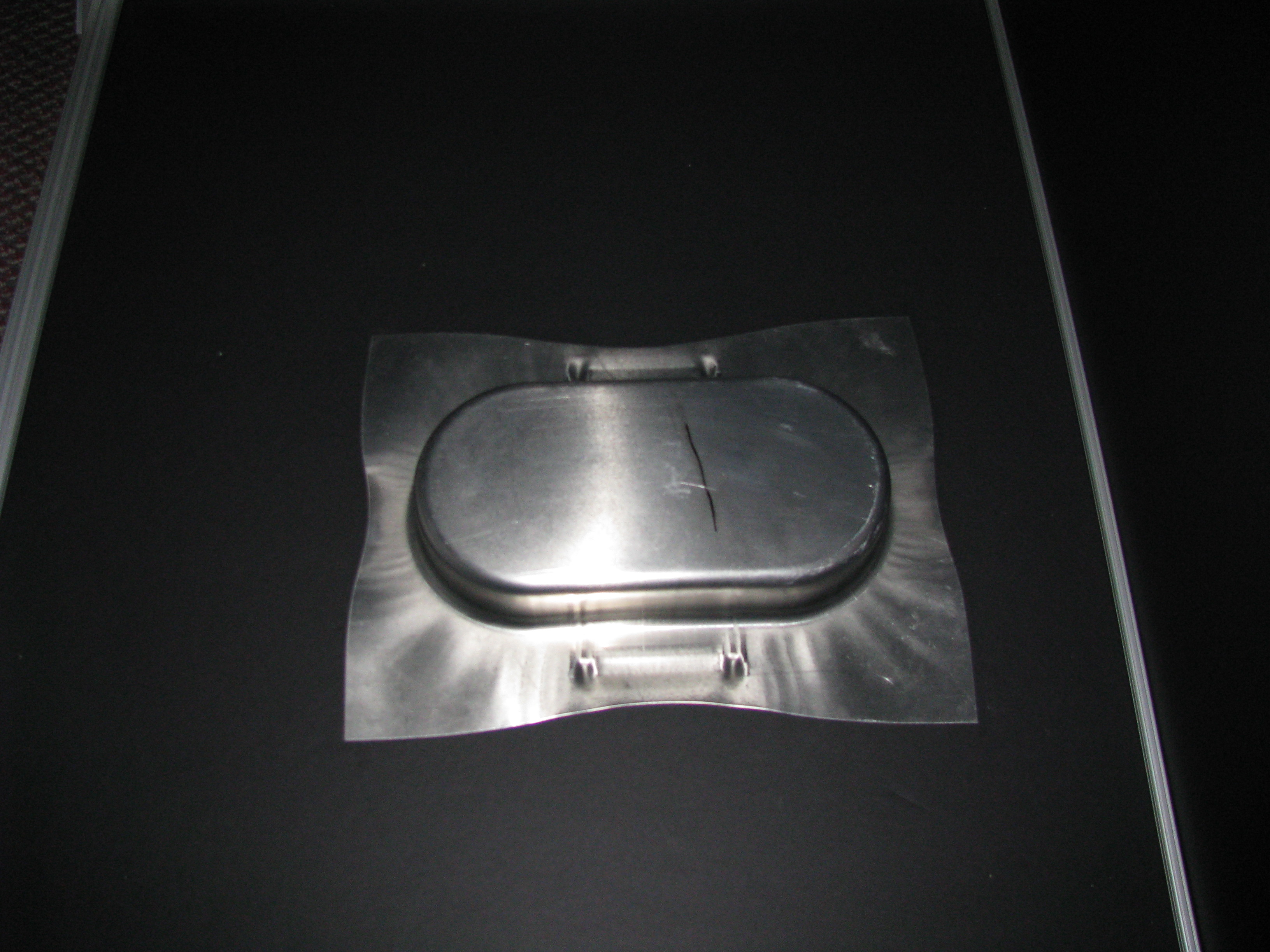}
\caption{Example defective image used to illustrate \texttt{CLIP}'s embedding process and zero-shot classification limitations.}
\label{fig:example_defect}
\end{figure}

To illustrate how \texttt{CLIP} operates in practice, consider the image in Figure \ref{fig:example_defect} showing a metallic component with a simulated defect. When this image (originally $3264\times2448$ pixels) is processed by the \texttt{ViT-L/14} model, it undergoes several transformations:
\begin{enumerate}[label=(\arabic*), nosep]
    \item The image is first preprocessed to the model's required input size ($336\times336$ pixels).
    \item The preprocessed image is embedded into a 768-dimensional feature vector as follows: $\mathbf{v} \in \mathbb{R}^{768} = [0.3713, 0.8574, -0.0626, 0.5874,  \ldots, -0.0016, -0.6489, -0.1530, -0.1975].$
    \item For zero-shot classification, we provide five text descriptions: (a) ``A defective metal component'', (b) ``A nominal metal component'', (c) ``An industrial part'', (d) ``A piece of sheet metal'', and (e) ``An artistic photograph''. Each description is first broken down into individual tokens (words and subwords) and padded to a fixed length of 77 tokens. This tokenization step transforms text into a format the model can process.
    \item Each tokenized description is then encoded into a 768-dimensional embedding. For example, the text ``A defective metal component'' is encoded into: \\$\mathbf{t} \in \mathbb{R}^{768} = [-0.0135, -0.0178, -0.0316, 0.0183,  \ldots, -0.0169, 0.0006, 0.0008, 0.0638].$
    \item The model computes cosine similarities between the normalized image embedding and each text embedding, which are then converted to probabilities using the \texttt{softmax} function.
\end{enumerate}

For this example, despite the visible defect (a black marker line simulating a crack), the following probabilities (extracted from the cosine similarities computed by the \texttt{CLIP} model) are presented to each of the pre-assigned five possible text descriptions/categories: 
\begin{enumerate}[nosep, label=(\alph*)]
    \item $Pr(\text{A defective metal component}) = 0.0616$
    \item $Pr(\text{A nominal metal component}) =  0.8506$
    \item $Pr(\text{An industrial part}) =  0.0495$
    \item $Pr(\text{A piece of sheet metal}) =  0.0380$
    \item $Pr(\text{An artistic photograph}) =  0.0001$
\end{enumerate}
Hence, \texttt{CLIP} would select ``A nominal metal component'' as the label since it had the highest similarity score and probability. 

This misclassification, where the model assigns the highest probability (0.8506) to ``A nominal metal component'' rather than ``A defective metal component'' (0.0616), illustrates a key limitation: while \texttt{CLIP} excels at general visual-textual understanding through its pretraining on 400 million image-text pairs, it may struggle with domain-specific tasks where the visual features differ significantly from its training distribution. Manufacturing defects, being relatively rare in general web images, represent exactly such a domain gap.

This limitation motivates our few-shot learning approach. Instead of relying on text descriptions alone, we provide \texttt{CLIP} with example images of both nominal and defective components. By computing embeddings for these examples, we create more reliable reference points in the embedding space. This allows \texttt{CLIP} to better distinguish between nominal and defective components based on visual similarity rather than text descriptions.

\section{Methods}
\label{sec:methods}

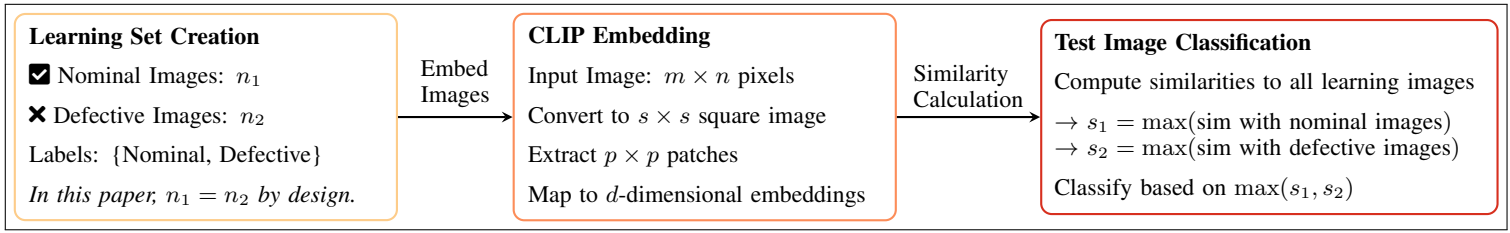
\begin{figure}[htb!]
\begin{adjustwidth}{-0.75in}{-0.75in}
\fbox{%
\centering
\begin{tikzpicture}[
    scale=1,
    transform shape,
    node distance=.6in,
    every node/.style={
        align=left,
        font=\footnotesize
    },
    header/.style={
        text centered,
        font=\bfseries
    },
    box/.style={
        thick,
        rounded corners,
        inner sep=5pt,
        fill=white,
        text width=1.85in,
        minimum width=2in
    },
    color2/.style={draw={rgb,255:red,253;green,204;blue,138}, fill=white},
    color3/.style={draw={rgb,255:red,252;green,141;blue,89}, fill=white},
    color4/.style={draw={rgb,255:red,215;green,48;blue,31}, fill=white},
    darkgreen/.style={draw=green!50!black, dashed, fill=white},
    arrow/.style={
        thick,
        ->,
        >=stealth
    }
]

\node[box, color2] (learning) {
    \textbf{Learning Set Creation}\\[5pt]
    \faCheckSquare \ Nominal Images: $n_1$\\[5pt]
    \faTimes \ Defective Images: $n_2$\\[5pt]
    Labels: \{Nominal, Defective\}\\[5pt]
    \textit{In this paper, $n_1 = n_2$ by design.}
};

\node[box, color3, right=.6in of learning] (clip) {
    \textbf{CLIP Embedding}\\[5pt]
    Input Image: $m \times n$ pixels\\[5pt]
    Convert to $s \times s$ square image\\[5pt]
    Extract $p \times p$ patches\\[5pt]
    Map to $d$-dimensional embeddings
};

\node[box, color4, right=.75in of clip, text width=2.25in] (classification) {
    \textbf{Test Image Classification}\\[5pt]
    Compute similarities to all learning images\\[5pt]
    $\rightarrow$ $s_1 = \max(\text{sim with nominal images})$\\
    $\rightarrow$ $s_2 = \max(\text{sim with defective images})$\\[5pt]
    Classify based on $\max(s_1, s_2)$
};


\draw[arrow] (learning) -- (clip) node[midway, above, sloped] {Embed\\Images};
\draw[arrow] (clip) -- (classification) node[midway, above, sloped] {Similarity\\Calculation};

\end{tikzpicture}%
}
\end{adjustwidth}
\caption{Workflow for learning and classification using \texttt{CLIP} embeddings for image quality control.}
\label{fig:workflow}
\end{figure}

Using \texttt{CLIP}'s visual encoder capabilities, our methodological framework introduces a deliberately simplified approach to few-shot learning in manufacturing quality control. While \texttt{CLIP} offers both visual and textual encoding pathways, we specifically use only its visual encoder for few-shot classification. We deliberately avoid text descriptions in the classification process. This design choice, though not leveraging \texttt{CLIP}'s full potential, establishes a minimal yet effective benchmark for the quality engineering community. More sophisticated approaches, such as \texttt{Tip-Adapter} \citep{zhang2022tip} or \texttt{Prompt-Generate-Cache} \citep{zhang2023prompt} methods, could potentially enhance performance by incorporating both visual and textual features. However, the collection of text captions by industrial machine vision systems is uncommon (which is why there is a need for image classification). Therefore, we prioritize establishing this more straightforward baseline for this expository paper. As illustrated in Figure \ref{fig:workflow}, our implementation consists of three primary components: learning set creation, \texttt{CLIP} embedding generation, and test image classification.

\subsection{Implementation Framework}

The learning set creation phase involves curating balanced learning and testing sets with equal numbers of nominal and defective examples ($n_1 = n_2$). This balanced design choice minimizes the impact of class imbalance \citep{megahed2021class}, allowing us to focus on \texttt{CLIP}'s inherent classification capabilities. Each image undergoes standardized preprocessing to meet \texttt{CLIP}'s input requirements through center cropping (preserving the central region of interest) and resizing to model-specific dimensions.

The \texttt{CLIP} embedding process transforms each preprocessed image into a high-dimensional feature vector. For instance, the \texttt{ViT-L/14} model maps each $336\times336$ pixel image into a 768-dimensional embedding space, while \texttt{ViT-B/32} produces 512-dimensional embeddings from $224\times224$ pixel inputs. This transformation occurs through \texttt{CLIP}'s patching mechanism, where images are divided into $p\times p$ patches ($p=14$ for \texttt{ViT-L/14}, $p=32$ for \texttt{ViT-B/32}) before being processed by the transformer architecture. Note that the patching idea is similar to the region of interest (ROI) idea in image monitoring applications; however, unlike typical implementations of ROIs \citep{megahed2012spatiotemporal,he2016image, okhrin2024control}, the patches are all equal in size and are non-overlapping.

The classification phase uses a maximum similarity approach that compares each test image against the learning set examples. Rather than relying on text descriptions, our few-shot implementation computes similarities between the test image's embedding and those of the learning set. The classification decision stems from identifying whether the highest similarity score corresponds to a nominal or defective example. This approach's computational complexity scales linearly with the learning set size during both the learning and testing phases.

\subsection{Rationale for the Selected Cases and Computational Experiment Overview}

We curated five case studies to systematically examine \texttt{CLIP}'s capabilities and limitations in manufacturing quality control.
\begin{enumerate}[label=(\arabic*), nosep]
    \item The first case study demonstrates \texttt{CLIP}'s effectiveness in applications where few-shot learning can work well with minimal modification, using metallic pan surface inspection. Here, we compare our approach against the results from \citet{megahed2012real}, showing that \texttt{CLIP} can outperform the original approach with easy-to-use code using the same experimental data.

    \item The second case study explores \texttt{CLIP}'s performance on lower-resolution extrusion profile images. We investigate how increasing the learning set size improves classification accuracy. The study examines the model's adaptability to images smaller than its standard $336\times336$ input size. This provides insights into practical applicability in resource-constrained environments.

    \item The third case study focuses on stochastic textured surfaces. We investigate how model choice impacts performance when defects are subtle or visually challenging to detect. The comparison between \texttt{ViT-L/14} and \texttt{ViT-B/32} variants demonstrates that model selection becomes crucial for nuanced defects requiring fine-grained feature detection.

    \item The fourth case study examines automotive assembly images. It illustrates the limitations of our simplified implementation when dealing with complex images containing multiple components. The study highlights scenarios requiring more sophisticated approaches. It provides guidance for practitioners considering alternative methods.

    \item The fifth case study examines \texttt{CLIP}'s application for automated classification of microstructural properties. We introduce this example to highlight how binary classification results can be seamlessly extended to multi-class outputs by leveraging the model's existing capabilities, without retraining or rerunning the model. This scalability demonstrates the practicality of using \texttt{CLIP} for numerous image-based quality engineering applications.
\end{enumerate}

\noindent Throughout these studies, we maintain consistent evaluation metrics, including accuracy, sensitivity/recall, specificity, precision, F1-score (harmonic mean of precision and recall), and AUC (area under the receiver operating characteristic curve), while adapting our experimental approach to address each case's unique challenges and objectives. We assume the reader is familiar with those commonly used classification metrics (otherwise, see \citet{lever2016classification} for a quick introduction). Our implementation emphasizes reproducibility through Python-based code that interfaces directly with \texttt{CLIP}'s package and OpenAI's API, requiring minimal dependencies beyond core requirements.

\subsection{Selected Benchmarks for our CLIP Implementation and their Rationale}

We compare our proposed CLIP-based few-shot learning approach against three references: (1) Published or Industry Benchmarks; (2) a MobileNetV2-based prototype classifier; and (3) Google Cloud Vertex AI's AutoML Vision.

\paragraph{Published and Industry Benchmarks.}
We report original metrics alongside our \texttt{CLIP} results when established baselines exist for a case study. These benchmarks measure progress relative to proven few-shot protocols and industrial best practices. This ensures meaningful and contextually relevant comparisons.

\paragraph{MobileNetV2 Few-Shot Prototype.}
We use MobileNetV2 as a lightweight baseline for few-shot image classification \citep{sandler2018mobilenetv2}. It is a well-established method in the literature (with over 29,000 Google Scholar citations) that preceded the release of the CLIP model.  MobileNetV2 uses depthwise-separable convolutions, inverted residual blocks, and linear bottlenecks to reduce parameters and computational complexity while retaining representational capacity. The model is pretrained on ImageNet \citep{deng2009imagenet} and provides robust feature embeddings that transfer readily to downstream tasks.

We freeze the entire MobileNetV2 backbone and extract per-image embeddings via its \texttt{features} trunk. We compute class prototypes by averaging the embeddings of the nominal and defective training samples. Test images are assigned to the nearest prototype using cosine similarity. The distance to the defective prototype serves as a continuous anomaly score.

\paragraph{AutoML Vision.}
We leverage AutoML Vision within Google Cloud's Vertex AI to benchmark against a turnkey commercial solution. This assesses the value proposition of a fully automated, managed service. AutoML automates the model development process by exploring architectures such as MnasNet (derived from MobileNetV2), EfficientNet, and other CNN variants \citep{vertexai-nas-overview2025, vertexai-searchspaces2025}. It applies automatic hyperparameter tuning to optimize performance for our task. AutoML incurs a cost of \$27.72 (USD) per case. Both the \texttt{CLIP}- and MobileNetV2-based prototypes require no paid computation.

We used the same test dataset for our \texttt{CLIP} based few-shot approach, MobileNetV2-based prototype classifier, and AutoML Vision to ensure fair evaluation. Vertex AI AutoML requires separate training, validation, and test datasets. We partitioned our original learning data by allocating 90\% for training and 10\% for validation. This follows Vertex AI's documentation \citep{vertexai-beginnersguide2025}. The validation set tunes the model's hyperparameters based on the model's performance. Our few-shot \texttt{CLIP}-based method and the MobileNetV2 prototype classifier require only learning and test sets. They involve no model optimization or hyperparameter tuning during training.

AutoML Vision operates on Google-managed infrastructure designed for scalable and high-performance training. The exact hardware is not disclosed. All few-shot learning experiments were conducted in a lightweight computing environment. We used a Google Colab instance with a 2-thread AMD EPYC virtual CPU. This disparity ensures that observed differences in training runtime or computational cost are not due to hardware advantages on our method's part.

\section{Case Study 1: Metallic Pan Surface}
\label{sec:case1}
\subsection{Dataset Description}

\begin{figure}[htb!]
\centering
\begin{subfigure}[b]{0.40\textwidth}
\centering
\includegraphics[width=.85\textwidth]{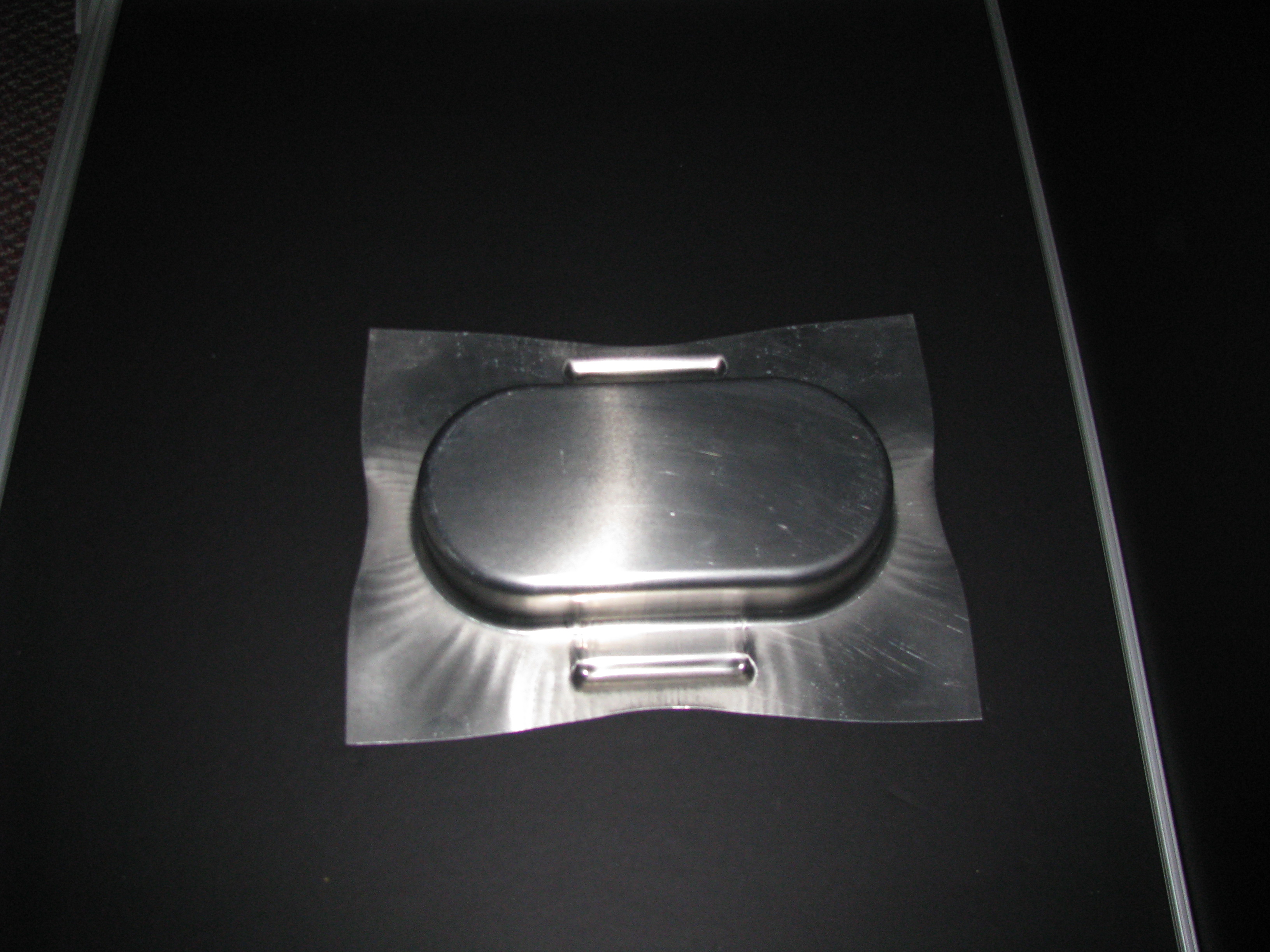}
\caption{Nominal pan}
\label{fig:pan_nominal}
\end{subfigure}
\hfill
\begin{subfigure}[b]{0.40\textwidth}
\centering
\includegraphics[width=.85\textwidth]{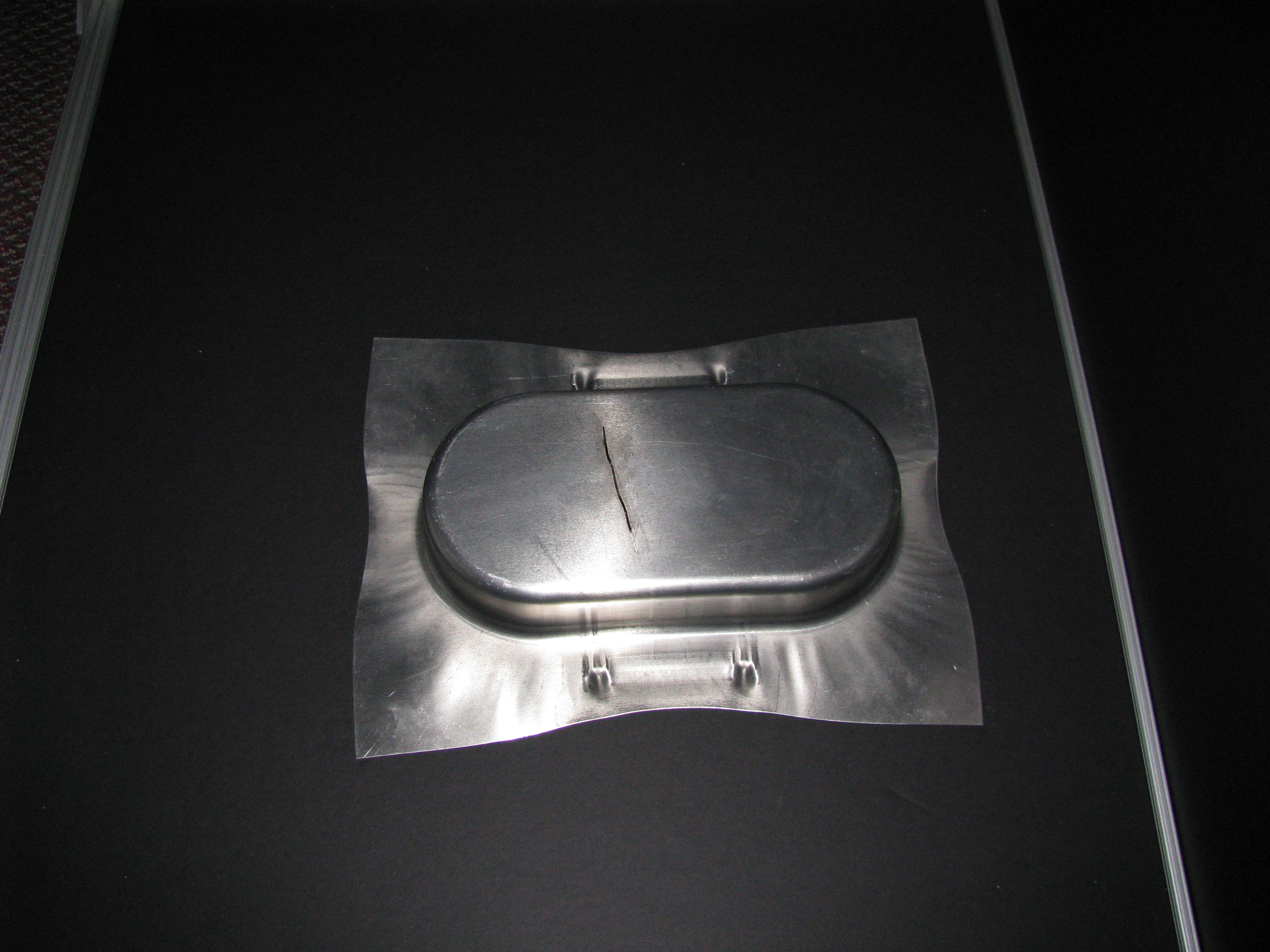}
\caption{Defective pan with a simulated crack}
\label{fig:pan_defective}
\end{subfigure}
\caption{Example images from the metallic pan surface inspection data: (a) shows a nominal pan surface without defects, while (b) displays a pan surface with a simulated crack marked in black.}
\label{fig:pan_examples}
\end{figure}

Our first case study examines the metallic pan surface inspection dataset from \citet{megahed2012real}. Figure \ref{fig:pan_examples} depicts representative nominal and defective images from the dataset, highlighting the nature of the simulated defects.

The original study captured images using a CANON SX 100 IS PowerShot 8.0 Mega-Pixels camera mounted on an aluminum frame ($3\mbox{ft}\times4\mbox{ft}\times3\mbox{ft}$). While conducted in a lab setting, the experimental design deliberately introduced variability to simulate shop floor conditions through:
\begin{enumerate}[label=(\arabic*), nosep]
    \item Part location variations:
    \begin{itemize}[nosep]
        \item Rotations around the central axis up to $\pm15^{\circ}$
        \item Translations up to $\pm0.6$ cm in both vertical and horizontal directions
        \item Variable lighting conditions across three different schemes
    \end{itemize}

    \item Defect characteristics:
    \begin{itemize}[nosep]
        \item Three fault categories: splits, cracks, and inclusions
        \item Crack and split widths ranging from 1.2 to 1.8 mm
        \item Defect lengths ranging from 1.8 to 4 cm
    \end{itemize}
\end{enumerate}

\noindent We adopted the original dataset split from the first author of \citet{megahed2012real}, using 10 nominal and 10 defective images for our few-shot learning. For the test set, we began with the 50 nominal and 50 defective images provided by \citet{megahed2012real}. Then, we removed training images, which yielded a test set of 39 nominal and 39 defective images. All images were never released publicly and hence, fall outside \texttt{CLIP}'s pre-training corpus.


\subsection{Zero-Shot Classification Performance}

Building upon our illustrative example in Section \ref{sec:background}, we expanded our zero-shot classification experiment in two ways. First, we simplified the classification to a binary problem (nominal vs. defective) rather than the five categories used in the illustration. Second, instead of analyzing a single image, we evaluated \texttt{CLIP}'s performance on the 78 test images (39 nominal, 39 defective). The use of multiple test images enables us to compute comprehensive classification metrics rather than just individual prediction probabilities (as we did in our illustrative example).

We provided \texttt{CLIP} with only textual descriptions: ``A metallic pan free of black scuff marks'' for the nominal class and ``A metallic pan with a simulated scuff mark drawn by a black marker'' for the defective class. Table \ref{tab:zeroshot_results} presents the classification metrics, revealing poor prediction performance.

\vspace{\baselineskip}

\begin{table}[htb!]
\centering
\caption{Zero-shot classification performance metrics for the metallic pan inspection case study.}
\vspace{-\baselineskip}
\label{tab:zeroshot_results}

\begin{tabular}{C{0.12\textwidth}C{0.12\textwidth}C{0.12\textwidth}C{0.12\textwidth}C{0.12\textwidth}C{0.12\textwidth}}
\toprule
\textbf{Accuracy} & \textbf{Sensitivity (Recall)} & \textbf{Specificity} & \textbf{Precision} & \textbf{F1-Score} & \textbf{AUC} \\
\midrule
0.615 & 0.231 & 1.000 & 1.000 & 0.375 & 0.559 \\
\bottomrule
\end{tabular}

\end{table}

The perfect specificity indicates that \texttt{CLIP} has correctly identified all nominal cases. Moreover, the perfect precision (1) suggests that it was always correct when \texttt{CLIP} identified a defect. However, the low sensitivity (0.231) indicates that \texttt{CLIP} missed many defective cases (i.e., classifying them as nominal) when relying solely on textual descriptions. This performance aligns with our earlier discussion about \texttt{CLIP}'s limitations in zero-shot manufacturing applications due to the domain gap between its training data and industrial applications.

\subsection{Few-Shot Classification Performance}

We then implemented our few-shot learning approach using the same 10 nominal and 10 defective images for learning as those used in \citet{megahed2012real}. Table \ref{tab:fewshot_results} displays these results, which demonstrate that our \texttt{CLIP}-based few-shot pipeline delivers exceptionally strong performance with a learning set of size only 20. With an overall accuracy of 0.91 and an AUC of 0.998, coupled with perfect specificity and precision, our approach eliminates false positives while maintaining a high true-positive rate (sensitivity = 0.82) and balanced F1-score of 0.9. Compared to the zero-shot baseline, the few-shot adaptation boosts recall more than threefold and roughly doubles the F1-score. Our few-shot learning adaptation of \texttt{CLIP} illustrates how even a small set of learning images can bridge \texttt{CLIP}'s domain gap and meet or exceed typical industrial inspection standards.


\begin{table}[ht]
\centering
\caption{Few-shot classification performance metrics for the metallic pan inspection case study.}
\vspace{-\baselineskip}
\label{tab:fewshot_results}

\begin{tabular}{@{}lcccccc@{}}
\toprule
\textbf{Method} & \textbf{Accuracy} & \textbf{Sensitivity} & \textbf{Specificity} & \textbf{Precision} & \textbf{F1-score} & \textbf{AUC} \\
\midrule
CLIP Few-Shot   & 0.910 & 0.821 & 1.000 & 1.000 & 0.901 & 0.998\\
\bottomrule
\end{tabular}

\end{table}

\vspace{-\baselineskip}


\subsection{Comparison with Other Benchmark Approaches}

Table \ref{tab:comparison1} compares our \texttt{CLIP}-based few-shot pipeline against three reference methods: the original benchmark from \cite{megahed2012real}, a MobileNetV2-based few-shot prototype (MobileNetV2-FS), and Google Cloud Vertex AI's AutoML Vision. Our approach achieves the highest accuracy (0.910). This improves on the original benchmark (0.880), MobileNetV2-FS (0.654), and AutoML (0.795). \texttt{CLIP}'s sensitivity more than doubles that of MobileNetV2-FS (0.308) and exceeds AutoML's 0.641. All methods maintain near-perfect specificity and precision.

Our pipeline runs in 9.1 minutes at zero monetary cost. Most of this time is spent loading the \texttt{CLIP} model and conducting the learning process, rather than classification, which occurs in near real-time. That being said, the MobileNetV2-FS approach only required 1.4 minutes for the entire process of few-shot learning and classification. On the other hand, the AutoML was significantly slower (93.0 minutes) and more expensive (\$27.72). Note that the original benchmark of \citet{megahed2012real} did not report AUC, runtime, or costs.

In this case study, our \texttt{CLIP}-based baseline outperforms both academic and commercial alternatives with minimal code and no fees. It preserves perfect false-positive control. Furthermore, it delivers practical runtime performance despite the model loading overhead.

\begin{table}[ht]
    \caption{Performance and runtime comparison for metallic pan surface inspection. Bolded values denote the best-performing model for each metric.}
    \label{tab:comparison1}
    \vspace{-\baselineskip}

    \begin{tabular}{lccccccr} \toprule
    \textbf{Model}	& \textbf{Accuracy}	& \textbf{Sensitivity}	& \textbf{Specificity}	& \textbf{Precision}	& \textbf{F1}	    & \textbf{AUC} & \textbf{Runtime (min)}	    \\ \midrule
    CLIP Few-Shot	& \textbf{0.910}	& 0.821	  & \textbf{1.000}	 & \textbf{1.000}	& \textbf{0.901}	  & \textbf{0.998}  & 9.1\\
    Benchmark*     & 0.880 & 0.760   & \textbf{1.000}    & \textbf{1.000}    & 0.857   & - -    & - - \\
    MobileNetV2-FS & 0.654 & 0.308   & \textbf{1.000}    & \textbf{1.000}    & 0.471   & 0.620  & \textbf{1.4}\\
    AutoML	        & 0.795	& 0.641	  & 0.949	 & 0.926	& 0.758	  & - -    & 93.0 \\
     \bottomrule
    \end{tabular}
    
\vspace{-0.5\baselineskip}
\footnotesize{*Results in \cite{megahed2012real} were based on 100 test images. The other approaches use 78 test images after removing any overlap with the training set. - - indicates the value is not available/reported.}
\end{table}

\vspace{-\baselineskip} 

\section{Case Study 2: 3D Printing Extrusion Profile Inspection}
\label{sec:case2}
\subsection{Dataset Description}

\begin{figure}[htb!]
\centering
\begin{subfigure}[b]{0.32\textwidth}
\centering
\includegraphics[width=85px]{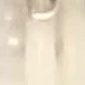}
\caption{Nominal extrusion profile}
\label{fig:extrusion_nominal}
\end{subfigure}
\hfill
\begin{subfigure}[b]{0.32\textwidth}
\centering
\includegraphics[width=85px]{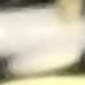}
\caption{Over-extruded profile}
\label{fig:extrusion_over}
\end{subfigure}
\hfill
\begin{subfigure}[b]{0.32\textwidth}
\centering
\includegraphics[width=85px]{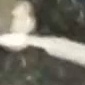}
\caption{Under-extruded profile}
\label{fig:extrusion_under}
\end{subfigure}
\caption{Examples from the extrusion profile dataset showing (a) a nominal profile with proper extrusion, (b) an over-extruded profile with excessive material flow, and (c) an under-extruded profile with insufficient material flow. Note: Images are shown at their original $85\times85$ pixel resolution.}
\label{fig:extrusion_examples}
\end{figure}

Our second case study examines the direct ink writing (DIW) 3D printing extrusion profile consistency inspection. Through the DIW printing process, complex devices made of composite materials can be obtained to realize different properties, such as high stretchability, heat insulation, and electrical conductivity. Guaranteeing the consistency of the process profile is critical to DIW. Figure \ref{fig:extrusion_examples} depicts representative nominal extrusion profile, over-extruded profile, and under-extruded profile images from the extrusion profile dataset.

The printer (\texttt{Ender-3}, Shenzhen Creality 3D Technology Co., Ltd.) is equipped with a custom fixture and an endoscope camera during the experiment. The syringe loaded with the ink is mounted on the fixture. In particular, the silica composite is used as the ink material. First, 3 grams of polyvinyl acetate (PVA) (360627, Sigma-Aldrich) is dissolved into 50 grams of deionized water, and the solution is heated under \SI{90}{\celsius} for 30 minutes with magnetic stirring. Then, 12.5 grams of silica nanoparticles (SkySpring Inc.) are added to the solution and mixed for 12 hours. Next, 0.35 grams of a viscosifier (Propylene carbonate, P52652, Sigma-Aldrich) are added into the silica solution and mixed for another two hours. Lastly, \SI{500}{\micro L} of 1-Octanol (A15977, Thermo Fisher Scientific Inc.) are added to remove the bubbles.

The motor pushes out the material and moves along the pre-defined route simultaneously. The endoscope camera monitors the printing condition in real-time, as shown in Figure \ref{fig:extrusion_examples}. We annotated the videos during printing and got the extrusion profile dataset. The data were not publicly released before and thus were not part of \texttt{CLIP}’s pre-training dataset.

\subsection{Zero-Shot Classification Performance}

We evaluated \texttt{CLIP}'s zero-shot classification performance using the following text prompts: ``An image of a normal extrusion profile from an additive manufacturing process'' for the nominal class and ``An image of an over-extruded or under-extruded profile from an additive manufacturing process'' for the defective class. Table \ref{tab:zeroshot_extrusion} presents the zero-shot classification results.

\vspace{\baselineskip}

\begin{table}[htb!]
\centering
\caption{Zero-shot classification performance metrics for the extrusion profile case study.}
\label{tab:zeroshot_extrusion}

\vspace{-\baselineskip}

\begin{tabular}{cccccc}
\toprule
\textbf{Accuracy} & \textbf{Sensitivity} & \textbf{Specificity} & \textbf{Precision} & \textbf{F1-Score} & \textbf{AUC} \\
\midrule
0.620 & 0.880 & 0.360 & 0.579 & 0.698 & 0.736 \\
\bottomrule
\end{tabular}
\end{table}

Unlike the metallic pan case study, where zero-shot classification showed high specificity but low sensitivity, here we observe the opposite pattern. The model achieved relatively high sensitivity (0.880) but poor specificity (0.360), suggesting a tendency to over-classify samples as defective. While zero-shot CLIP resulted in a moderate accuracy of 0.620, it cannot be used for this application in practice since it would produce too many false alarms.

\subsection{Few-Shot Learning and the Impact of Learning Set Size}

We first evaluated \texttt{CLIP}'s few-shot learning performance using a baseline set of 20 examples (10 nominal, 10 defective). This setup was selected since it is similar to what we did in the previous case study. Table \ref{tab:fewshot_extrusion} presents three initial results, where we can see that the predictive performance is poor. We suspect that this is mainly due to the low resolution images available in the case study; the extrusion profile images were only $85\times85$ pixels, significantly smaller than \texttt{CLIP}'s expected input dimensions ($336\times336$ for the \texttt{ViT-L/14} encoder model). Note that throughout this case study, the number of nominal images was equal to the number of defective images (which was equally split between over- and under-extruded defects). Hereafter, we use the term ``per class'' to denote the nominal and overarching defective classes.

\vspace{\baselineskip}

\begin{table}[htb!]
\centering
\caption{Few-shot classification performance metrics for the extrusion profile case study using 10 examples per class.}
\label{tab:fewshot_extrusion}

\vspace{-\baselineskip}
\begin{tabular}{@{}lcccccc@{}}
\toprule
\textbf{Method} & \textbf{Accuracy} & \textbf{Sensitivity} & \textbf{Specificity} & \textbf{Precision} & \textbf{F1-score} & \textbf{AUC} \\
\midrule
CLIP Few-Shot   & 0.570 & 0.400 & 0.740 & 0.600 & 0.480 & 0.620 \\
\bottomrule
\end{tabular}

\end{table}

\begin{figure}[htb!]
\centering
\begin{adjustwidth}{-0.85in}{-0.85in}
\includegraphics[width=1.25\textwidth, frame, trim={0 0.5in 0 0.5in}, clip]{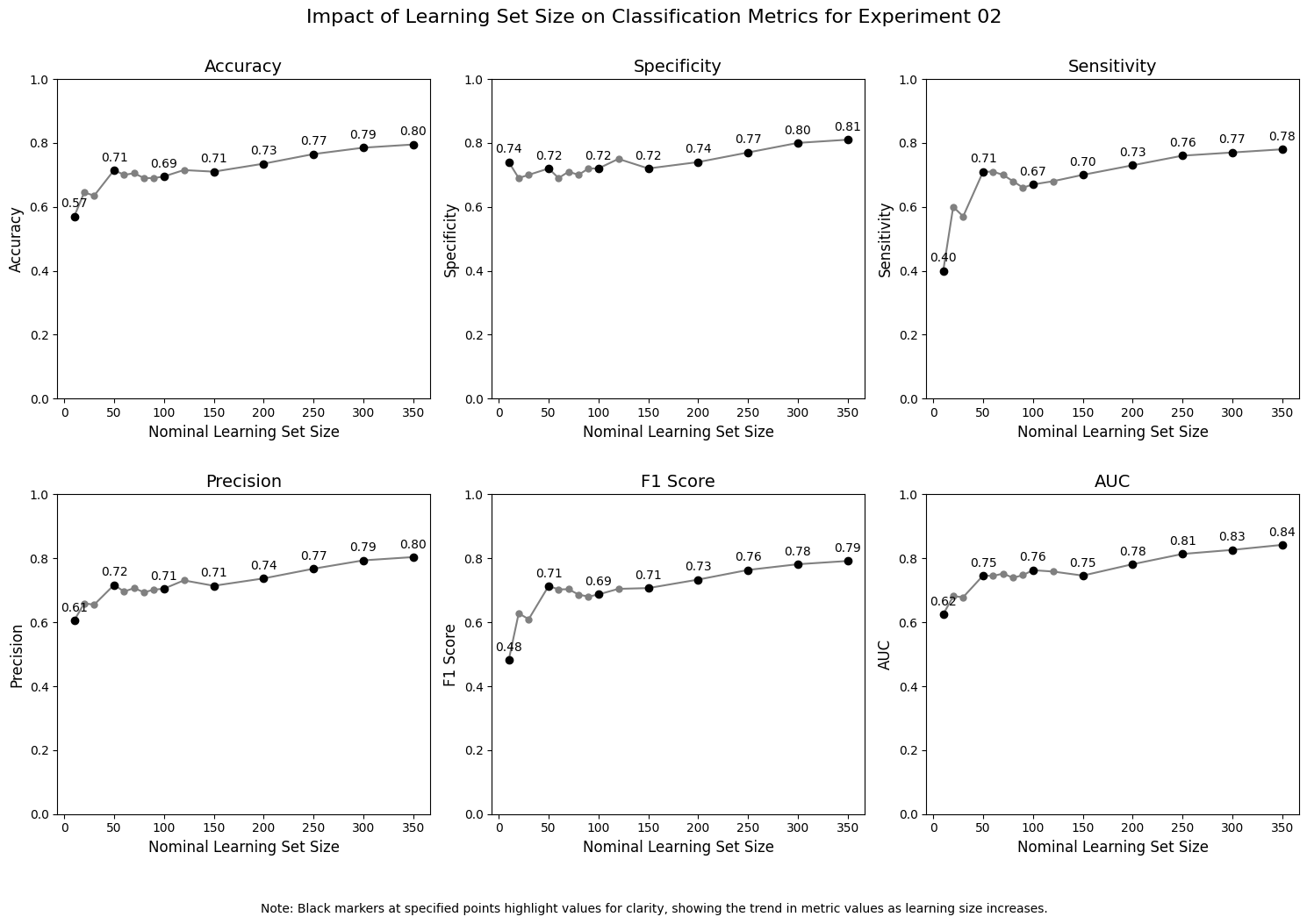}
\end{adjustwidth}
\caption{Impact of learning set size on classification metrics for the extrusion profile case study. Black markers highlight specific measurements at key per class sample sizes (10, 50, 100, 150, 200, 250, 300, and 350 nominal examples).}
\label{fig:learning_curve_extrusion}
\end{figure}

To investigate whether additional examples could improve the baseline results, we conducted a comprehensive analysis varying the learning set size from 10 to 350 nominal examples (i.e., total learning sample size is double these numbers) while maintaining a fixed test set for consistent comparison. Our results, depicted in Figure \ref{fig:learning_curve_extrusion}, revealed three key findings. First, even with just 10 examples per class, the model achieved an accuracy of 0.57 with  40\% of the defects (sensitivity of 0.4) and 74\% of the nominal cases (specificity of 0.74) correctly identified. Second, we observed improved performance between 10 and 100 per class examples for some classification metrics. For example, accuracy improved from 0.57 to 0.71, and AUC increased from 0.62 to 0.75; however, specificity remained relatively stable. Finally, performance improvements plateaued beyond 200 per class examples. The maximum values for our predictive metrics were achieved at 350 nominal learning examples, with an accuracy of 0.80, specificity of 0.81, and sensitivity of 0.78.

These findings are notable given that the $85\times85$ pixel resolution falls well below \texttt{CLIP}'s designed input specifications. Despite requiring significant upscaling, the model achieved performance gains with modest data requirements. The improvement with just 100 per class examples contrasts with traditional deep learning approaches, which typically require thousands of training samples.

The results suggest that while CLIP can effectively adapt to lower-resolution inputs through few-shot learning, optimal performance likely requires images closer to the model's intended dimensions. This limitation warrants consideration when applying this approach to manufacturing applications with low-resolution imaging systems.

\subsection{Comparison with Other Benchmark Approaches}

For this case study, no published results are available. We therefore compare only two methods against our approach under the same few-shot training regime (100 labeled images equally split among classes). AutoML is trained on 90 images, with 10 reserved for validation. The MobileNetV2-based few-shot prototype and our \texttt{CLIP}-based few-shot pipeline both use all 100 images. Table \ref{tab:comparison2} captures the results of the comparison. MobileNetV2-FS attains the highest accuracy (0.805), specificity (0.890), precision (0.867), F1-score (0.787), and AUC (0783). MobileNetV2-FS leads in raw performance. \texttt{CLIP}-based method delivers reasonable sensitivity and discriminative power (AUC = 0.746) at zero cost and with minimal code, but it lags behind MobileNetV2-FS on overall accuracy, precision, and F1.

\begin{table}[ht]
    \caption{Performance and runtime comparison for 3D printing extrusion profile inspection. Bolded values denote the best-performing model for each metric.}
    \label{tab:comparison2}
    \vspace{-\baselineskip}

    \begin{tabular}{lccccccr} \toprule
     \textbf{Model}	& \textbf{Accuracy}	& \textbf{Sensitivity}	& \textbf{Specificity}	& \textbf{Precision}	& \textbf{F1}	    & \textbf{AUC}& \textbf{Runtime (min)}	\\ \midrule
         CLIP Few-Shot	& 0.700	 & 0.710  & 0.690  & 0.696	& 0.703	& 0.746 & 17.0 \\
         MobileNetV2-FS & \textbf{0.805}  & 0.720  & \textbf{0.890}  & \textbf{0.867}  & \textbf{0.787} & \textbf{0.783} & \textbf{1.3} \\
         AutoML	        & 0.775	 & \textbf{0.800}  & 0.750  & 0.762	& 0.781	& - -   & 103.0
	    \\ \bottomrule
    \end{tabular}
    
    \vspace{-0.5\baselineskip}
\footnotesize{- - indicates the value is not available/reported.}
\end{table}

\vspace{-\baselineskip} 

\section{Case Study 3: Stochastic Textured Surfaces}
\label{sec:case3}

\subsection{Dataset Description}

Our third case study examines stochastic textured surfaces (STS), a unique class of manufacturing quality control data that presents distinct challenges compared to traditional profile monitoring. STS data include woven textiles, surface metrology data from machined, cast, or formed metal components, and microscopy images of material microstructures \citep{bui2018monitoring}. As noted by \citet{bui2018monitoring}, STS data present several challenges: (a) unlike our first case study of metallic surfaces where the gold standard is simply a non-textured surface of constant intensity or cases with deterministically repeated patterns where computer-aided design models can serve as references, STS has no definitive ``gold standard'' image; (b) under normal process conditions, there exist infinitely many valid images that exhibit identical statistical properties while being completely different at the pixel level; and (c) these images cannot be easily aligned, transformed, or warped into a standard reference image due to their inherent stochasticity. These challenges render both traditional profile monitoring and image inspection/monitoring approaches inapplicable \citep{megahed2011review,bui2018monitoring}. This fundamental limitation of conventional approaches motivates our investigation of \texttt{CLIP}'s potential for STS image inspection, i.e., can vision transformers (with few-shot examples) help alleviate the aforementioned limitation of traditional approaches?

\begin{figure}[htb!]
\centering
\begin{subfigure}[b]{0.32\textwidth}
\centering
\includegraphics[width=\textwidth]{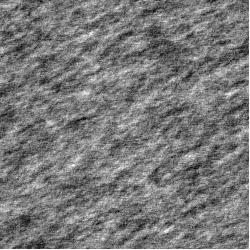}
\caption{Nominal texture}
\label{fig:sts_nominal}
\end{subfigure}
\hfill
\begin{subfigure}[b]{0.32\textwidth}
\centering
\includegraphics[width=\textwidth]{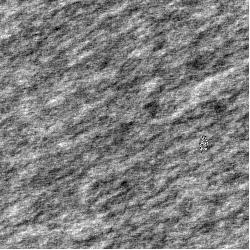}
\caption{Local defect}
\label{fig:sts_local}
\end{subfigure}
\hfill
\begin{subfigure}[b]{0.32\textwidth}
\centering
\includegraphics[width=\textwidth]{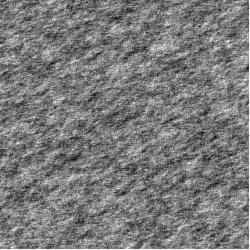}
\caption{Global defect}
\label{fig:sts_global}
\end{subfigure}
\caption{Examples of simulated stochastic textured surfaces (STS): (a) nominal texture with spatial autoregressive parameters $\phi_1=0.6$ and $\phi_2=0.35$, (b) texture with a localized defect disrupting the pattern, and (c) texture with a global defect simulated by reducing both parameters by 5\%.}
\label{fig:sts_examples}
\end{figure}

To systematically evaluate \texttt{CLIP}'s performance on STS image classification, we utilized the \texttt{spc4sts} package \citep{bui2021spc4sts} to generate a comprehensive dataset of simulated textile images. This approach enabled the controlled generation of a large-scale dataset, including both nominal and defective samples with known characteristics, as well as the precise manipulation of defect types (local vs. global) and magnitudes. Our experimental dataset included 1,000 nominal images representing standard weave patterns, 500 images with local defects characterized by localized disruptions in the weave pattern, and 500 images with global defects involving systematic shifts in weave parameters. Each image was generated at $250\times250$ pixels following \citet{bui2021spc4sts}'s recommended parameters. Nominal images were generated using spatial autoregressive parameters $\phi_1=0.6$ and $\phi_2=0.35$. Local defects were imposed using the package's defect generation function, and global defects were simulated by reducing both parameters by 5\%. Figure \ref{fig:sts_examples} shows representative examples of the nominal and two defect categories. Note the subtle visual differences between the cases.

We used simulated images because \citet{bui2018monitoring} only provided six defect images, which would make it challenging to effectively split the data into few-shot learning and testing sets. We provide the \faRProject \, code (with a fixed seed), and the generated images on our GitHub repository to facilitate the reproduction of our image generation process and/or analysis.

\subsection{Zero-Shot Classification Performance}

We used the following two captions to evaluate \texttt{CLIP}'s zero-shot classification performance: ``An image of a textile material with consistent weave patterns, showing no visible defects or irregularities'' for the nominal class; and ``A woven textile, featuring a visible tear, defect, or a disruption in the material's weave structure'' for the defective class. Table \ref{tab:zeroshot_sts} presents these results, revealing significant limitations in \texttt{CLIP}'s zero-shot capabilities for STS inspection. The perfect specificity but zero sensitivity indicates that \texttt{CLIP} classified all images as nominal, i.e., failing to detect any defects. This performance is particularly poor compared to our previous case studies, likely due to the subtle nature of STS defects and their significant deviation from \texttt{CLIP}'s training data distribution.

\vspace{\baselineskip}

\begin{table}[htb!]
\centering
\caption{Zero-shot classification performance metrics for the STS case study.}
\label{tab:zeroshot_sts}

\vspace{-\baselineskip}

\begin{tabular}{cccccc}
\toprule
\textbf{Accuracy} & \textbf{Sensitivity} & \textbf{Specificity} & \textbf{Precision} & \textbf{F1-Score} & \textbf{AUC} \\
\midrule
0.500 & 0.000 & 1.000 & 0.000 & 0.000 & 0.171 \\
\bottomrule
\end{tabular}
\end{table}

\subsection{Model Selection Impact on Few-Shot Learning}

\begin{figure}[htb!]
\centering
\begin{adjustwidth}{-0.85in}{-0.85in}
\includegraphics[width=1.25\textwidth, frame, trim={0 0.5in 0 0.5in}, clip]{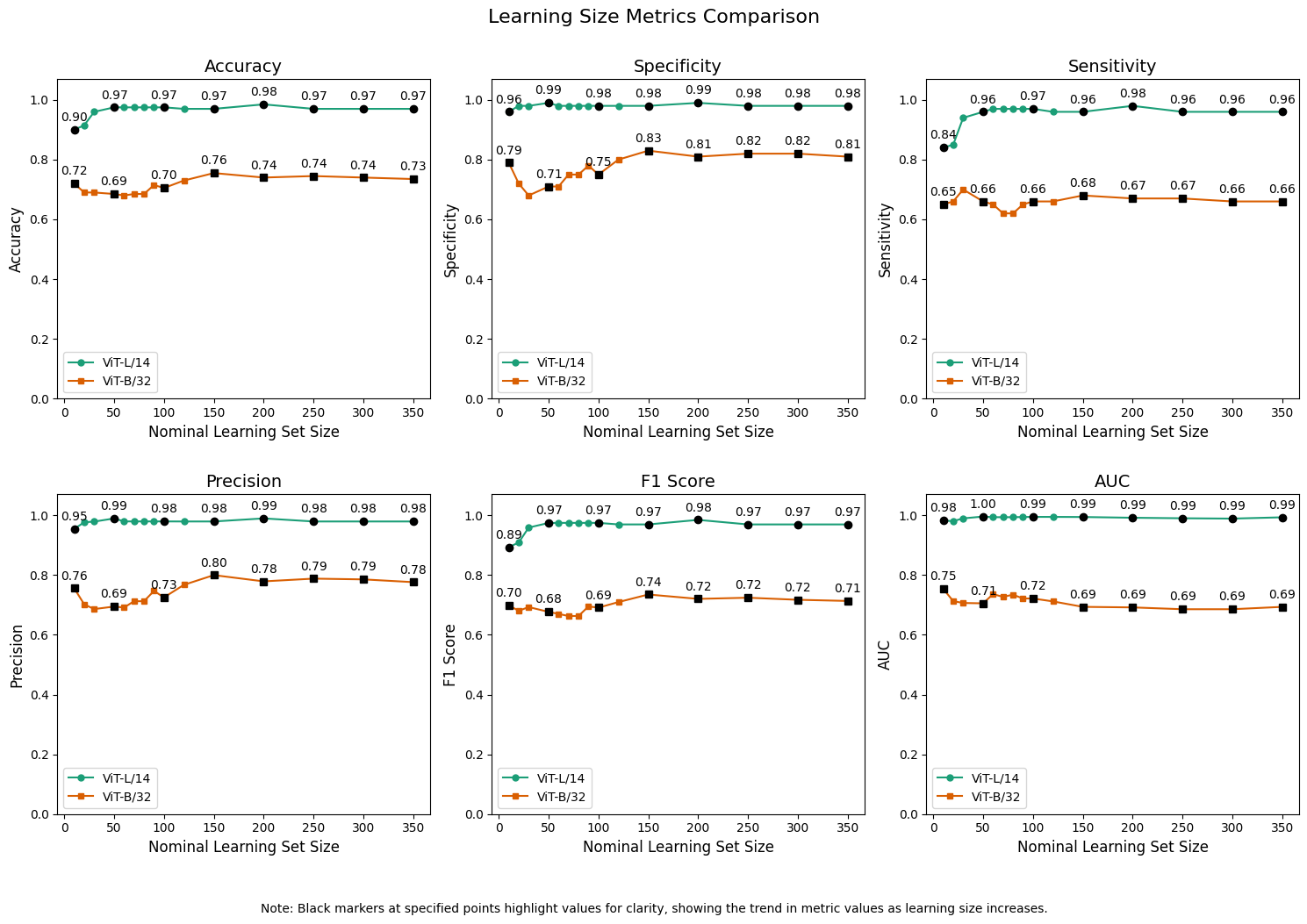}
\end{adjustwidth}
\caption{Impact of encoder model and learning set size on classification metrics for the STS case study. Black markers highlight specific measurements at key sample sizes (10, 50, 100, 150, 200, 250, 300, and 350 nominal examples).}
\label{fig:model_comparison_sts}
\end{figure}

We suspect that one contributor to the poor performance of the zero-shot approach is the complexity of the STS images. To investigate whether a more detailed vectorization of the images might improve performance, we compare two image encoders: \texttt{ViT-L/14} and \texttt{ViT-B/32}. The \texttt{ViT-L/14} encoder uses a $14 \times 14$ pixel patch size, giving a more fine-tuned vectorization of the images than the \texttt{ViT-B/32}, which uses a $32 \times 32$ pixel patch.   Figure \ref{fig:model_comparison_sts} shows how classification metrics evolved with increasing learning set size for both models. Note that throughout this case study, the number of nominal images was equal to the number of defective images (which was equally split between local and global defects).

The results reveal several key findings. The \texttt{ViT-L/14} model demonstrated superior performance across all metrics, achieving an impressive accuracy of 0.97 with just 50 nominal and 50 defective learning examples. Its performance remained consistently high, exceeding 0.95 for all metrics once the learning set size surpassed 100 nominal and 100 defective examples. It also achieved a near-perfect AUC of 0.99 across most of the learning range. In contrast, the \texttt{ViT-B/32} model exhibited significantly weaker performance. It had a maximum accuracy of only 0.76, even when provided with 350 nominal and 350 defective examples. Its improvement with increased learning set size was inconsistent. The \texttt{ViT-B/32} achieved markedly lower sensitivity (0.66) and precision (0.78). The performance gap between the models persisted even at the maximum learning set size, with \texttt{ViT-L/14} outperforming \texttt{ViT-B/32} by 24 percentage points in accuracy (0.97 vs. 0.73), 30 percentage points in sensitivity (0.96 vs. 0.66), and 17 percentage points in specificity (0.98 vs. 0.81).

These findings highlight the importance of encoder model selection for STS inspection tasks. The superior performance of \texttt{ViT-L/14} suggests that its larger model capacity and finer-grained patch size ($14\times14$ vs $32\times32$) allow for capturing the subtle patterns that characterize STS defects. This finding aligns with theoretical expectations, as smaller patch sizes should better preserve the local spatial correlations that define STS patterns. While the input images ($250\times250$ pixels) were smaller than \texttt{ViT-L/14}'s designed input size ($336\times336$ pixels), the model still performed well. This finding suggests that the model's architecture is robust to moderate downscaling when the resolution remains sufficient to capture relevant texture patterns.

The results also demonstrate that the \texttt{ViT-L/14} \texttt{CLIP}-based few-shot learning model can detect local and global STS defects without requiring explicit statistical modeling of the underlying stochastic process. This finding is particularly significant as it suggests that \texttt{ViT-L/14} can maintain high classification performance even with smaller-than-designed input image sizes, provided they retain sufficient resolution to capture the textural characteristics. This capability, combined with the elimination of the need for robust parameter estimation and monitoring scheme design, represents a significant practical advantage over traditional approaches to STS inspection.

\subsection{Comparison with Other Benchmark Approaches}

In the third case study on STS inspection (Table \ref{tab:comparison3}), the original training set of 100 images was used in all three approaches. \texttt{CLIP}-based model achieves the highest overall accuracy (0.975), F1-score (0.975), specificity (0.990), and AUC (0.996). Furthermore, its sensitivity of 0.960 is quite competitive with the highest sensitivity achieved (0.970) using the AutoML approach.  For this case study, our zero-cost approach offers the best trade-off between accuracy, runtime, and implementation simplicity compared to the two reference approaches.

As a contextual reference, \citet{bui2018monitoring} utilized a control-chart based method for STS images; thus, they reported the average power (i.e., sensitivity) between 0.858 and 1.000 at a fixed Type I error of $\alpha=0.003$ (a controlled specificity of 0.997). While our method and theirs have utilized different test sets and evaluation criteria, these figures underscore that our \texttt{CLIP}-based approach attained sensitivity and specificity comparable to those of established statistical inspection techniques, which further support its viability as a lightweight, few-shot baseline for texture-based quality control.

\begin{table}[ht]
    \caption{Performance and runtime comparison for stochastic textured surfaces.  Bolded values denote the best-performing model for each metric.}
    \label{tab:comparison3}
    \vspace{-\baselineskip}

    \begin{tabular}{lccccccr} \toprule
    \textbf{Model}	& \textbf{Accuracy}	& \textbf{Sensitivity}	& \textbf{Specificity}	& \textbf{Precision}	& \textbf{F1}	    & \textbf{AUC} & \textbf{Runtime {min}}	    \\ \midrule
    CLIP Few-Shot	 & \textbf{0.975}  & 0.960	& \textbf{0.990}	  & \textbf{0.990}	& \textbf{0.975}	 & \textbf{0.996}  & 17.1 \\
    MobileNetV2-FS  & 0.725  & 0.510   & 0.940   & 0.895   & 0.650  & 0.583  & \textbf{1.3} \\
    AutoML	         & 0.880  & \textbf{0.970}	& 0.790	  & 0.822	& 0.890	 & - -    & 203.0 \\ \bottomrule
    \end{tabular}
    
\vspace{-0.5\baselineskip}
\footnotesize{- - indicates the value is not available/reported.}
\end{table}

\vspace{-0.5\baselineskip} 

\section{Case Study 4: Renault's Pipe Staple}
\label{sec:case4}

\subsection{Dataset Description}

\begin{figure}[htb!]
\centering
\begin{subfigure}[b]{0.45\textwidth}
\centering
\includegraphics[width=0.9\textwidth]{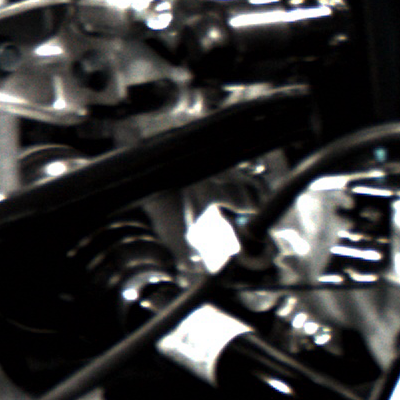}
\caption{Nominal assembly}
\label{fig:staple_nominal}
\end{subfigure}
\hfill
\begin{subfigure}[b]{0.45\textwidth}
\centering
\includegraphics[width=0.9\textwidth]{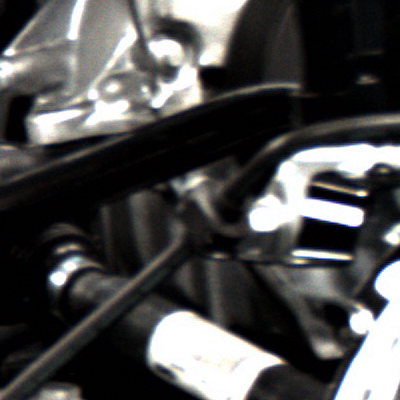}
\caption{Assembly with missing clamp}
\label{fig:staple_defective}
\end{subfigure}
\caption{Examples from the pipe staple dataset: (a) shows a properly secured cable with the metallic clamp in place, while (b) displays a defective assembly where the securing clamp is missing.}
\label{fig:staple_examples}
\end{figure}

Our fourth case study examines the recently published Renault pipe staple dataset \citep{carvalho2024detecting}, which presents a real-world challenge in automotive assembly inspection. The dataset consists of images capturing the assembly of flexible cables secured by metallic rectangular clamps in the automotive manufacturing process. Figure \ref{fig:staple_examples} shows representative examples of nominal and defective cases from the dataset.

This dataset is particularly interesting for our study for several reasons. First, published in July 2023 (approximately two years after \texttt{CLIP}'s development), these images were definitively not part of \texttt{CLIP}'s training data. Second, the dataset was originally constructed for unsupervised classification research, with their test dataset containing both nominal and defective examples. Given that their datasheet indicates the data were collected in chronological order, we leveraged this temporal structure in our study. Specifically, we used the first 50 nominal and 50 defective images from their test dataset for learning, reserving the final 50 images of each class for testing. This chronological splitting approach simulates a realistic implementation scenario where earlier examples inform the classification of later cases.

The inspection task in this dataset presents several unique challenges \citep{carvalho2024detecting}. The images capture varying camera angles and perspectives of the assembled components, contain complex backgrounds with multiple automotive parts visible, and exhibit real manufacturing environment lighting variations and shadows. These characteristics make the visual detection of missing clamps particularly challenging, as the differences between properly secured and unsecured cables can be quite subtle depending on the viewing angle and lighting conditions.

\subsection{Zero-Shot Classification Performance}

For zero-shot classification, we evaluated \texttt{CLIP} using the following two prompts: ``A close-up of an automotive assembly process showing a flexible cable secured by a metallic rectangular clamp with rounded edges'' for the nominal class, and ``A close-up of an automotive assembly process focusing on a flexible cable, where the clamp that typically secures the pipe during assembly is notably absent'' for the defective class. Table \ref{tab:zeroshot_staple} presents the zero-shot classification results.

\vspace{\baselineskip}

\begin{table}[htb!]
\centering
\caption{Zero-shot classification performance metrics for the pipe staple case study.}
\label{tab:zeroshot_staple}

\vspace{-\baselineskip}

\begin{tabular}{cccccc}
\toprule
\textbf{Accuracy} & \textbf{Sensitivity} & \textbf{Specificity} & \textbf{Precision} & \textbf{F1-Score} & \textbf{AUC} \\
\midrule
0.500 & 1.000 & 0.000 & 0.500 & 0.667 & 0.710 \\
\bottomrule
\end{tabular}
\end{table}

The zero-shot classification results revealed significant limitations in \texttt{CLIP}'s ability to handle this complex inspection task without example images. The perfect sensitivity but zero specificity indicates that \texttt{CLIP} classified all images as defective, suggesting an inability to distinguish the visual cues that differentiate properly secured cables from those missing clamps.

\subsection{Few-Shot Classification Performance}

Given the dataset's size limitations and our chronological splitting approach, we implemented few-shot learning using a fixed few-shot example set of 50 nominal and 50 defective examples from the earlier portion of the dataset. Table \ref{tab:fewshot_staple} presents these results.

\vspace{\baselineskip}

\begin{table}[htb!]
\centering
\caption{Few-shot classification performance metrics for the pipe staple case study using 50 examples per class.}
\label{tab:fewshot_staple}

\vspace{-\baselineskip}

\begin{tabular}{cccccc}
\toprule
\textbf{Accuracy} & \textbf{Sensitivity} & \textbf{Specificity} & \textbf{Precision} & \textbf{F1-Score} & \textbf{AUC} \\
\midrule
0.580 & 0.560 & 0.600 & 0.583 & 0.571 & 0.608 \\
\bottomrule
\end{tabular}
\end{table}

While these results show modest improvement over zero-shot classification, they fall short of the performance levels achieved in our previous case studies and state-of-the-art approaches. For example, \citet{carvalho2024detecting} demonstrated that specialized computer vision methods can achieve mean AUC values as high as 98.9\% ($\pm$0.6\%) on this dataset. This substantial performance gap suggests that our simplified \texttt{CLIP}-based approach, while effective for some inspection tasks, may be insufficient for complex assembly inspection scenarios involving multiple components, varying viewpoints, and subtle defect characteristics. The results reinforce that some manufacturing inspection tasks require more sophisticated approaches, particularly when dealing with real-world assembly operations where defect detection involves complex spatial relationships between multiple components.

\subsection{Comparison with Other Benchmark Approaches}

Our \texttt{CLIP}- and MobileNetV2 few-shot models each used all 100 images for learning. AutoML was trained on 90 images with 10 held out for validation. Table \ref{tab:comparison4} shows that AutoML attains perfect performance across every metric. This reflects its strength when classes are easily separable visually \citep{vertexai-beginnersguide2025}. The MobileNetV2-FS prototype achieves 0.750 accuracy, 0.720 sensitivity, 0.780 specificity, and an F1-score of 0.742 in just 1.37 min at zero cost. Both methods outperform our \texttt{CLIP}-based few-shot approach, which posts 0.580 accuracy, 0.560 sensitivity, 0.600 specificity, and an F1-score of 0.571 in 13.2 min.

\begin{table}[ht]
    \caption{Performance and runtime comparison for Renault's pipe staple. Bolded values denote the best-performing model for each metric.}
    \label{tab:comparison4}
    \vspace{-\baselineskip}

    \begin{tabular}{lccccccr} \toprule
    \textbf{Model}	& \textbf{Accuracy}	& \textbf{Sensitivity}	& \textbf{Specificity}	& \textbf{Precision}	& \textbf{F1}	& \textbf{AUC} & \textbf{Runtime (min)}	\\ \midrule
    \textbf{CLIP Few-Shot}	& 0.580	  & 0.560  & 0.600	& 0.583	 & 0.571  &	\textbf{0.608} & 13.2 \\
    \textbf{MobileNetV2-FS} & 0.750   & 0.720  & 0.780  & 0.766  & 0.742  & 0.590 & \textbf{1.4}\\
    \textbf{AutoML}	        & \textbf{1}   & \textbf{1}  & \textbf{1}  & \textbf{1}	 & \textbf{1}  & - -   & 98.0 \\
 \bottomrule
    \end{tabular}
    
\vspace{-0.5\baselineskip}
\footnotesize{- - indicates the value is not available/reported.}
\end{table} 

\section{Case Study 5: Microstructure Images}
\label{sec:case5}
\subsection{Dataset Description}

Manufacturing applications with stringent functional requirements demand consistent, repeatable microstructural properties across parts. Recent advancements in metal additive manufacturing have demonstrated that microstructure can be easily tuned by altering the scanning strategy, offering significant flexibility in microstructure customization \citep{liu2022additive}.
  Current inspection methods rely on destructive testing and Electron Backscattered Diffraction (EBSD) imaging \citep{bostanabad2018computational} to assess material properties at the microstructural level.

Manual evaluation by human experts introduces significant limitations in industrial practice \citep{larmuseau2021race}. The growing volume of material characterization data makes manual labeling a resource-intensive process. Expert analysis inherently contains subjective biases, which can limit the detection of subtle variations within natural data variance. Therefore, the materials science and manufacturing communities have increasingly adopted automated classification methods for microstructure analysis \citep{gola2018advanced, pratap2022machine}. Our investigation of this case is motivated by the suitability of our proposed few-shot learning approach with \texttt{CLIP} to applications with limited training data. If successful, our proposed approach can automate the inspection of microstructure images to reduce the cost associated with expert-driven microstructural measurements.

\begin{figure}[htb!]
    \centering
    \includegraphics[width=0.87\textwidth]{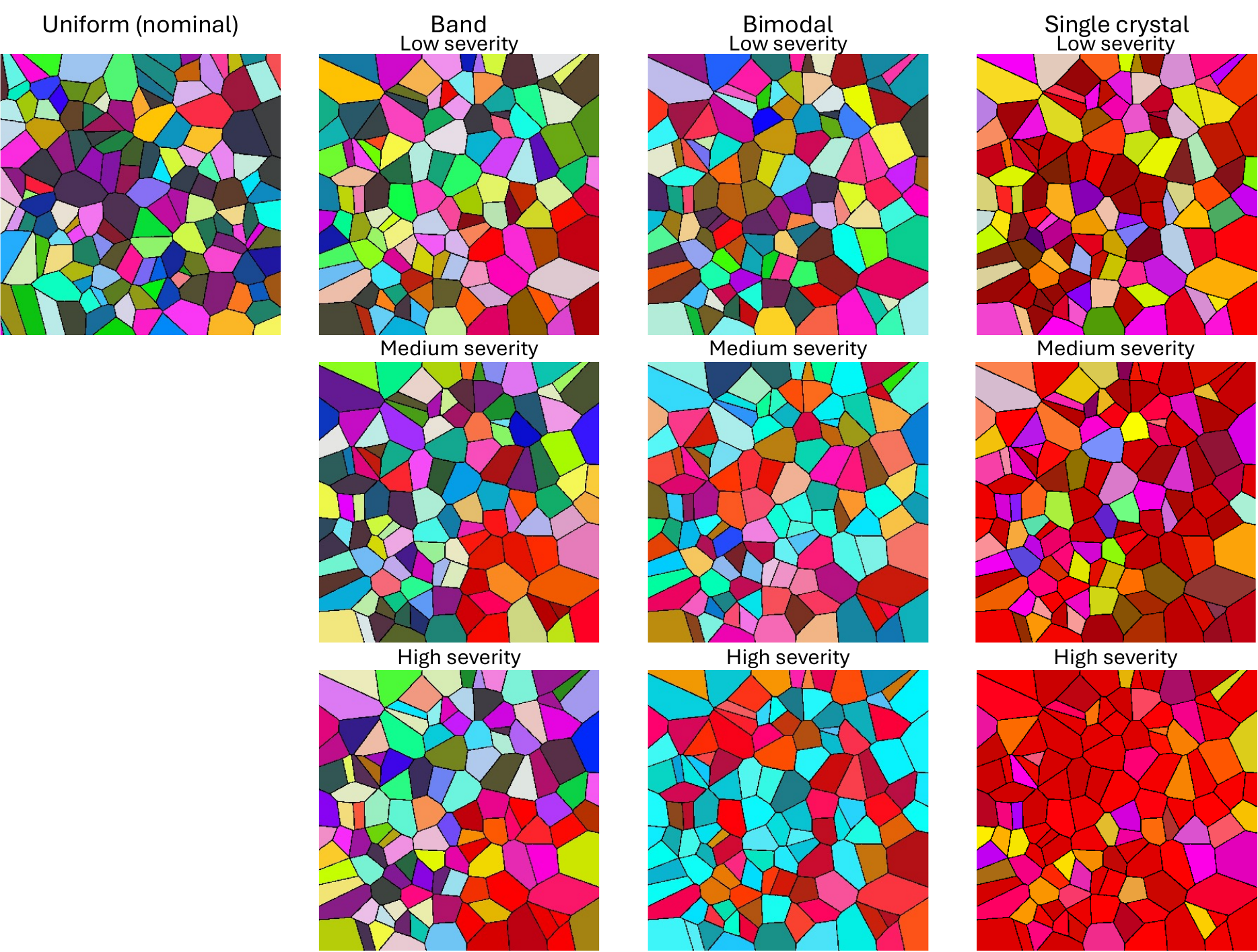}
    \caption{Examples of simulated microstructures with different crystal orientation distributions}
    \label{fig:microstructures}
\end{figure}

We simulated EBSD microstructure images by combining random Voronoi tessellation with crystal orientation distribution functions (ODFs) \citep{barrett2019generalized}. Voronoi edges represent grain size and morphology. The ODF assigns crystal orientations to grains based on predefined probability functions. Crystal orientations map to colors using standard EBSD measurement schemes. Figure \ref{fig:microstructures} illustrates the four example simulated microstructures investigated in our study: (a) \ul{uniform (nominal)} with random crystal orientations from a uniform distribution, (b) clustered orientation showing a \ul{band} of grains with predominant crystal orientation surrounded by uniform ODF grains, (c) \ul{bimodal} with crystal orientations from ``bimodal'' distribution, and (d) single crystal-like where crystals align toward one orientation. These classes reflect EBSD patterns in metal additive manufacturing. The uniform ODF represents nominal conditions. The other defective classes indicate defects resulting from variations in local or global heating and cooling dynamics during the manufacturing process.

Our ODF simulations used weighted schemes to generate three severity levels per class (``low'', ``medium'', ``high''). These levels represent different degrees of deviation from the nominal uniform ODF. The simulated microstructural images are 400 x 400 pixels in size. For technical implementation details, we refer the interested reader to \cite{wei2025lowdimensional}.

\subsection{Few-Shot Classification Performance}

Our study used 72 simulated nominal images for few-shot learning, with eight images for each defective class--severity combination ($3 \text{ classes [band, bimodal, single-crystal] } \times 3 \text{ severity levels}$). The test data consisted of 225 images for the nominal class and 25 images for each defect-severity level combination. The training and testing datasets were balanced based on the overarching nominal and defective classes, i.e., the total number of nominal images was equal to the total number of defective images (irrespective of their category and severity). Furthermore, the number of images within each primary defective class (``band'', ``bimodal'', and ``single-crystal'') was equal. Similarly, the number of images within the sub-labels of the primary defective class and severity level combinations was equal. For the sake of conciseness, we only report the few-shot classification performance to highlight the potential of using \texttt{CLIP} with few-shot learning in multiclass image classification applications.

Similar to case studies 1-4, Table \ref{tab:fewshot_microstructure_binary} presents the performance of the few-shot learning for the binary image classification experiment (nominal vs. defective). Here, we define an image as defective if it belongs to any of the three primary defective classes (``band'', ``bimodal'', and ``single-crystal''). The few-shot learning binary classification results show that the model excelled at detecting defects with a sensitivity of 0.987, i.e., 222 out of the 225 defective images were correctly classified. Furthermore, nominal images were correctly classified as nominal 80.9\% of the time, with 182 out of 225 nominal images classified correctly. Due to the balanced nature of the dataset, the overall accuracy represents the average of those two numbers and is approximately 90\%.

\vspace{0.5\baselineskip}

\begin{table}[htb!]
\centering
\caption{Few-shot classification performance metric for the microstructure classification case study.}

\vspace{-0.75\baselineskip}

\begin{tabular}{@{}lcccccc@{}}
\toprule
\textbf{Test mode} & \textbf{Accuracy} & \textbf{Sensitivity} & \textbf{Specificity} & \textbf{Precision} & \textbf{F1-score} & \textbf{AUC} \\
\midrule
Two-class   & 0.898 & 0.987 & 0.809 & 0.838 & 0.906 & 0.952 \\
\bottomrule
\end{tabular}
\label{tab:fewshot_microstructure_binary}
\end{table}

\vspace{-0.5\baselineskip}

\begin{figure}[htb!]
\captionsetup[subfigure]{justification=centering}
\centering
\begin{subfigure}[t]{\textwidth}
    \centering
    \begin{tcolorbox}[colback=white, colframe=gray!90, title=\texttt{Reformatted CSV-Line Output for Image microstructure\_039.png}, fontupper=\small, fonttitle=\small, width =0.9\textwidth]
\begin{verbatim}
datetime_of_operation: 2025-01-15T17:04:03.471179,

num_few_shot_nominal_imgs: 72,

image_name: microstructure_039.png,

classification_result: Defective,

non_defect_prob: 0.495,

defect_prob: 0.505,

nominal_description: Image microstructure_234.png:
  An microstructure image with uniformly distributed grain sizes and colors.,

defective_description: Image microstructure_001.png:
  An microstructure image with a low single crystal structure.
\end{verbatim}
\end{tcolorbox}
\caption{Our implementation's reformatted output for test image ``microstructure\_039.png''.}
\label{fig:csv_output_reformatted}
\end{subfigure}

\vspace{1em}

\begin{subfigure}[t]{0.28\textwidth}
    \centering
    \includegraphics[width=.95\textwidth]{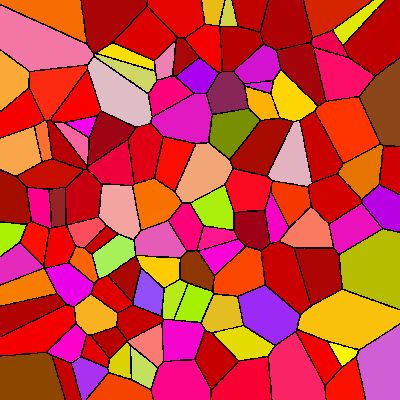}
    \caption{Test image\\ \texttt{microstructure\_039.png}}
\end{subfigure}
\begin{subfigure}[t]{0.28\textwidth}
    \centering
    \includegraphics[width = .95\textwidth]{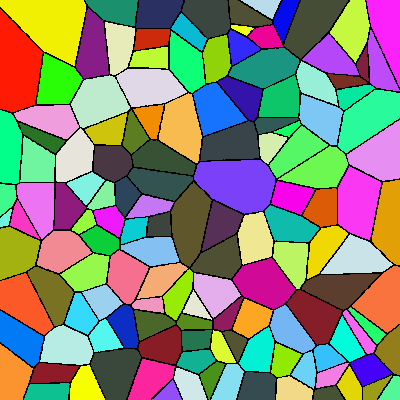}
    \caption{Nominal learning image\\ \quad \texttt{microstructure\_234.png}}
\end{subfigure}
\begin{subfigure}[t]{0.28\textwidth}
    \centering
    \includegraphics[width=.95\textwidth]{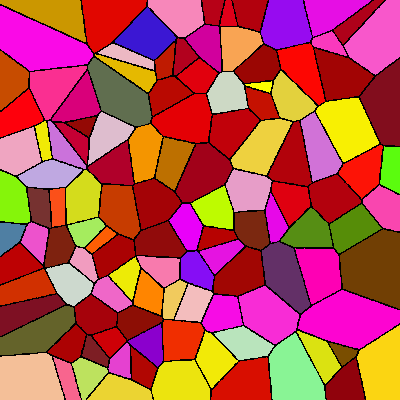}
    \caption{Defective learning image \texttt{microstructure\_001.png}}
\end{subfigure}
\caption{(a) shows the reformatted output for the test image \texttt{microstructure\_039.png}, including its classification as defective. It is more similar to \texttt{microstructure\_001.png}, whose corresponding description includes ``low single-crystal structure.'' (b) presents the test image \texttt{microstructure\_039.png}. (c) displays the most similar nominal image, \texttt{microstructure\_234.png}, identified by our model. (d) shows the most similar defective image, \texttt{microstructure\_001.png}, identified by our model. The binary classification result of defective is correct. In our four-class classification problem (nominal, ``band'', ``bimodal'', and ``single-crystal''), the model correctly classifies our test image as ``single-crystal.'' However, in our 10-class classification (nominal + nine defect types/severity combinations) problem, the classification of ``low single-crystal'' is incorrect since the true label of \texttt{microstructure\_234.png} is ``medium single-crystal.'' Note that we do not need to rerun the model for the 4-class and 10-class classification problems.}

\end{figure}

Given the relatively large number of potential defect classes (up to nine if we consider the class-severity combination), it would be interesting to investigate the model's multiclass classification performance. Note that we can extract its multiclass classification performance without needing to ``retrain'' or ``rerun'' the model. Recall that our approach takes every test image and computes its cosine similarity with every image within the ``nominal'' and ``defective'' training dataset. From this comparison, we identify the most relevant image within each nominal and defective class, its match probability, and training-based captions (detailed image description) for its most similar training ``nominal'' and ``defective'' images. Consider our test set image \texttt{microstructure\_039.png} to explain this concept further. The output from our \texttt{CLIP-}based implementation is shown in Figure \ref{fig:csv_output_reformatted}. The Figure shows that the overall classification result is defective, as the probability of being defective (0.505) is slightly larger than that of being non-defective (0.495). We can also see that its most similar defective image is \texttt{microstructure\_001.png}, which had a description/caption that ends with ``low single crystal structure''. We extracted this description using simple text extraction strategies to obtain our multiclass results. We utilized two extraction strategies: (a) focusing only on whether it is ``band'', ``bimodal'', or ``single-crystal'', and (b) assigning it to one of the nine defective sub-labels shown in Figure \ref{fig:microstructures} by extracting severity (``low'', ``medium'', and ``high'') and the primary defect class (``band'', ``bimodal'', and ``single-crystal'') indicators.

\begin{figure}[htbp]
    \centering
    \begin{subfigure}[b]{0.67\textwidth}
        \centering
        \includegraphics[width=0.98\textwidth]{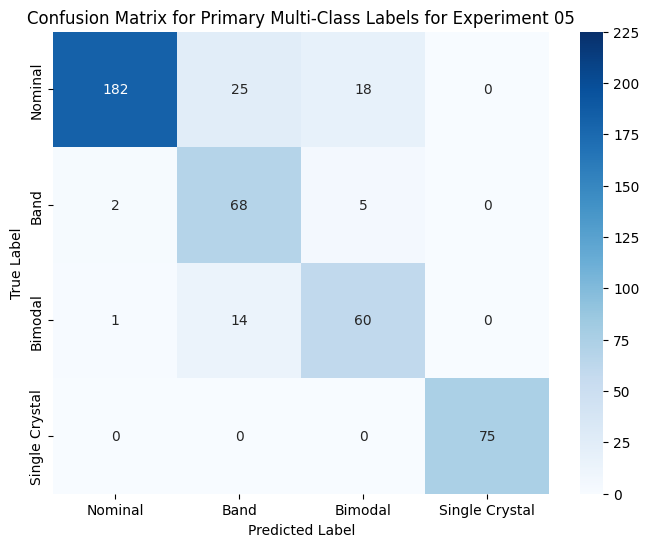}
        \caption{Primary Confusion Matrix.}
        \label{fig:primary_confusion_matrix}
    \end{subfigure}%
    \begin{subfigure}[b]{0.33\textwidth}
        \centering
        \begin{tabular}{lc}
            \toprule
            Metric                   & Value  \\ \midrule
            Accuracy                 & 0.856  \\
            Sensitivity (Macro Avg)       & 0.879  \\
            Precision (Macro Avg)    & 0.836  \\
            F1-Score (Macro Avg)     & 0.849  \\ \bottomrule
        \end{tabular}
        \caption{Primary Metrics.}
        \label{tab:primary_metrics}
    \end{subfigure}

    \vspace{1em} 

    \begin{subfigure}[b]{0.67\textwidth}
        \centering
        \includegraphics[width=0.98\textwidth]{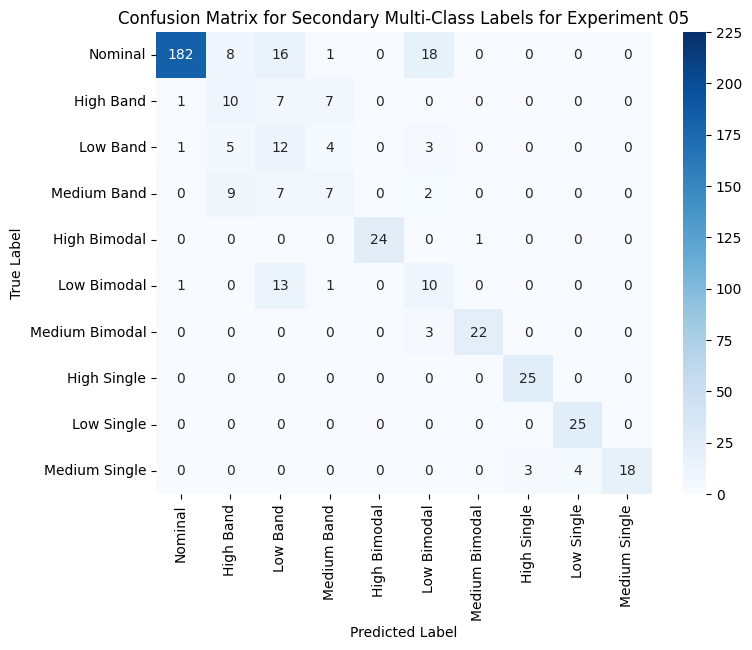}
        \caption{Secondary Label (Sublabel) Confusion Matrix.}
        \label{fig:secondary_confusion_matrix}
    \end{subfigure}%
    \begin{subfigure}[b]{0.33\textwidth}
        \centering
        \begin{tabular}{lc}
            \toprule
            Metric                   & Value  \\ \midrule
            Accuracy                 & 0.744  \\
            Sensitivity (Macro Avg)       & 0.693  \\
            Precision (Macro Avg)    & 0.685  \\
            F1-Score (Macro Avg)     & 0.678  \\ \bottomrule
        \end{tabular}
        \caption{Sublabel Metrics.}
        \label{tab:secondary_metrics}
    \end{subfigure}

    \caption{Confusion matrices and the corresponding classification performance metrics for primary and secondary multiclass labels for Case Study 5.}
    \label{fig:combined_figure}
\end{figure}

Figure \ref{fig:combined_figure} captures the performance of our \texttt{CLIP} model for both extraction approaches. The first row captures the confusion matrix for the primary labels (nominal, ``band'', ``bimodal'', and ``single-crystal'') alongside its corresponding classification metrics table. The confusion matrix highlights the distribution of actual and predicted labels, with correct classifications appearing on the diagonal and off-diagonal values representing specific classification errors. Unsurprisingly, Figure \ref{fig:primary_confusion_matrix} shows that 182 out of 225 nominal cases were correctly classified as nominal (precisely equal to the specificity values reported in Table \ref{tab:fewshot_microstructure_binary}). This is expected since we did not retrain or rerun the model. However, the extraction of labels allows us to see how the 43 misclassified nominal images were labeled, with 25 labeled as ``band'' and 18 labeled as ``bimodal''. Furthermore, we can see that all 75 (100\% of) ``single-crystal'' defective images were classified correctly. We can also see that 68 out of 75 ($\simeq 91$\%) ``band'' defects and 60 out of 75 (80\%) ``bimodal'' defects were correctly classified. The accompanying metrics table (Figure \ref{tab:primary_metrics}) shows an overall classification accuracy of 85.6\%, and reports the values for several macro-averaged metrics.

To help explain the idea of ``macro-average'' computation, let us consider our reported macro-averaged sensitivity of 87.9\%. This is computed by performing a one-vs-all comparison for each class. For each class, sensitivity is calculated as the ratio of true positives (diagonal values) to the sum of true positives and false negatives (row total minus the diagonal). The sensitivity for the nominal class is
$\frac{182}{182 + 25 + 18} = 0.809,$ and we have shown that sensitivities for ``band'', ``bimodal'', and ``single-crystal'' are $0.907, \ 0.8, \text{ and } 1,$ respectively. Hence, the macro-averaged sensitivity is computed by averaging these values across all classes: $\frac{0.809 + 0.907 + 0.8 + 1.0}{4} = 0.879$ \citep{grandini2020metrics}. This computational approach was applied to the remaining metrics shown in Figure \ref{tab:primary_metrics}.

The second row in Figure \ref{fig:combined_figure} captures the confusion matrix for the nominal class and the nine defective sublabels alongside their corresponding classification metrics table. As expected, with the increase in the classification degree of difficulty (the model must correctly identify both the primary defect class and its severity), the overall accuracy drops to 74.4\%. However, compared to the primary classification example, the decrease in performance can only be attributed to misclassification errors among different severities of the same type of defect (crystal orientation) rather than errors between different crystal orientations.

A closer investigation of the results, reported in Figure \ref{fig:combined_figure}, also reveals that a higher misclassification error was obtained for the ``bimodal'' ODF compared to the other classes. It was misclassified mainly as ``band''. This makes practical sense since the ``band'' and ``bimodal'' distributions involve a ``bimodal'' orientation. The difference in labeling arises from the spatial distribution of orientations: in the ``band'' class, orientations are spatially clustered, whereas in the ``bimodal'' class, they are randomly distributed.

These results highlight the proposed approach's capability to classify complex microstructure data through few-shot classification. The results also show the simplicity of extending our binary classification approach to multiclass classification without rerunning the model. This allows practitioners to perform classifications sequentially: first distinguishing between nominal and defective data and then identifying specific types of defects. Note that our results for this case study will likely be improved if we increase the number of learning examples, as shown in case studies 2 and 3. However, we do not explore this here for three reasons: (a) the obtained results demonstrate the approach's potential and were satisfactory; (b) our primary rationale for presenting this case study was to show the ease of translating the binary classification results into a multiclass classification output, without the need to rerun or ``retrain'' the model; and (c) conciseness.

\subsection{Comparison with Other Benchmark Approaches}

A standard industrial practice for quantitative analysis and classification of microstructure images relies on computing simple synthetic descriptors. Typical descriptors relate to salient characteristics like grain size, morphology, and crystal orientation. We use two synthetic descriptors as a benchmark: (a) the entropy of RGB pixel intensities, and (b) the dominant color fraction. Their choice was motivated by the nature of simulated microstructures, which differ in terms of their Orientation Distribution Functions (ODFs). The RGB levels vary while the average grain and morphology distributions remain constant. These two descriptors capture microstructural heterogeneity (entropy) and the prevalence of any single crystallographic variant (dominant fraction) in two compact numbers. These are critical for texture quantification, grain statistics, and subsequent structure–property correlations \citep{Brandon&Kaplan2013, WeietAl2025}.

Each RGB microstructure was first converted to a grey-scale image. The distribution of grey levels was sampled with 256 equal‐width bins to form a probability mass function, $p_k$. The Shannon entropy was then estimated as:
\[
H \;=\; -\sum_{k} p_{k}\,\log p_{k}
\]
For the dominant color fraction, all pixel colors were reshaped into an $N\times3$ matrix in RGB space and partitioned into $K=5$ clusters using the $k‐$means algorithm (with three random initializations, a maximum of 100 iterations). Being $N_j$ the number of pixels assigned to cluster $j$, the dominant‐color fraction was defined as:
\[
f_{\mathrm{dom}}
\;=\;
\frac{\displaystyle \max_{1 \le j \le K} N_{j}}{N}.
\]
\noindent
The two descriptors were computed for every image. A k-NN classifier was trained and tested in the resulting bivariate space with $K=5$. The microstructure dataset was divided into training and test sets, analogously to the \texttt{CLIP} approach. This allowed us to evaluate performance against a benchmark representative of common industrial practice for microstructure image characterization and classification.

Table \ref{tab:comparison5} captures the binary classification performance of our \texttt{CLIP-}based approach along with the three reference methods: (a) the common industrial practice of computing simple synthetic descriptors, (b) the MobileNetV2-based few-shot prototype, and (c) AutoML Vision. AutoML achieves the highest raw performance with 0.967 accuracy and 0.966. The descriptor-based benchmark trails substantially at 0.653 accuracy and 0.578 F1. Our few-shot \texttt{CLIP} approach and the MobileNetV2-based few-shot prototype outperform these industry-standard descriptors. They are competitive with the AutoML approach (especially since they incur zero costs and offer much shorter runtimes).

\texttt{CLIP}-based few-shot offers a unique advantage: its outputs support interpretable and hierarchical classification. Our implementation returns descriptive prompts for both nominal and defective reference images, as shown in Figure \ref{fig:csv_output_reformatted}. These descriptions are generated without retraining the model. They enable practitioners to extend the binary classification setup to multiclass tasks using simple text parsing and matching techniques. The capacity for sequential classification without retraining or modifying the model underscores the flexibility of our few-shot approach for applications involving complex and nuanced visual structures.

\begin{table}[ht]
    \caption{Performance and runtime comparison for microstructure images. Bolded values denote the best-performing model for each metric.}
    \label{tab:comparison5}
    \vspace{-\baselineskip}

    \begin{tabular}{lccccccr} \toprule
    \textbf{Model}	& \textbf{Accuracy}	& \textbf{Sensitivity}	& \textbf{Specificity}	& \textbf{Precision}	& \textbf{F1}	& \textbf{AUC}  & \textbf{Runtime (min)}	   \\ \midrule
    \textbf{CLIP Few-Shot}	     & 0.898 & 0.987  & 0.809 & 0.838 & 0.906 & \textbf{0.952} & 34.4 \\
    \textbf{Benchmark*}          & 0.653 & 0.632  & 0.667 & 0.533 & 0.578 & 0.728 & - - \\
    \textbf{MobileNet-FS}        & 0.887 & 0.773  & \textbf{1.000} & \textbf{1.000} & 0.872 & 0.854 & \textbf{2.7} \\
    \textbf{AutoML}	             & \textbf{0.967} & \textbf{0.991}  & 0.942 & 0.991 & \textbf{0.966} & - -   & 103.0 \\
 \bottomrule
    \end{tabular}

    \vspace{-0.5\baselineskip}
    \footnotesize{*Industrial Practice. - - indicates the value is not available/reported.}
\end{table} 

\section{Discussion}
\label{sec:disc}
\subsection{Insights from the Experiments}

Our case studies demonstrate that \texttt{CLIP}, when adapted for few-shot learning, can serve as a powerful yet simple baseline for image-based quality control in a variety of contexts. Specifically, our \texttt{CLIP} applications demonstrate:
\begin{itemize}
    \item \textbf{Strong Performance with Minimal Data:} Most of our applications of the \texttt{CLIP} model achieved high accuracy with small learning sets--91\% accuracy with 10 examples per class (metallic pan case), 97\% accuracy with 50 examples per class (STS case), 90\% accuracy with 72 examples (microarray structure case), and 80\% accuracy with 350 examples per class in the low resolution ($85\times85$ pixel) extrusion case.  These are well below the $10,000+$ examples used in deep learning applications \citep[e.g., see][]{kang2024modern}.
    \item \textbf{Competitive Performance with Implementation Advantages:} \texttt{CLIP} demonstrated strong performance across diverse manufacturing scenarios with notable practical benefits. Performance varied by application complexity: it outperformed all methods in two case studies (metallic pan and STS), exceeded industry standards and slightly outperformed MobileNetV2-FS in microstructure classification, delivered competitive discriminative power (AUC 0.746 vs. 0.783) relative to MobileNetV2-FS in extrusion profile inspection (despite the below specification image size), but was significantly outperformed by AutoML in the complex pipe staple assembly task. Beyond predictive performance, CLIP offers substantial implementation advantages. It provides major advantages over AutoML in speed (9.1-34.4 minutes vs. 93.0-203.0 minutes), cost (zero vs. \$27.72 per case), user control, and interpretability. This combination positions CLIP as an attractive first baseline for practitioners seeking rapid deployment of image-based quality control systems.
    \item \textbf{A Nonlinear Relationship between Learning Size and Performance:} When increasing the learning sample size, some performance metrics improved rapidly for the first 50-100 examples per class with diminishing returns beyond 150-200 examples.  The specific inflection point of the nonlinear improvement varied by application. In addition, performance gains across the size of the learning set were nonmonotonic.  This phenomenon is likely due to the addition of learning set images, with some additional learning cases making it easier (or harder) to classify test images.
    \item \textbf{Robustness to Image Resolution:} \texttt{CLIP} successfully processed images ranging from $250\times250$ pixels (STS case) to full-resolution $3264\times2448$ images (metallic pan case). While the \texttt{ViT-L/14} model is designed for $336\times336$ pixel inputs, it can effectively handle smaller and larger images through appropriate preprocessing. The impact of resolution on performance appears application-dependent, with optimal results generally achieved when image dimensions are larger or relatively close to the model's design specifications. In the case of lower resolution images ($85\times85$ pixels in the extrusion case), a larger training sample (350 images per class) was required to achieve adequate performance.
    \item \textbf{Sensitivity to Image Complexity:} The model excelled with single-component images (metallic pan). Notably, \texttt{CLIP} performed well on STS images, a traditionally challenging domain for automated inspection. However, the \texttt{CLIP} model did not perform well with multi-component scenarios (pipe staple case).  Our results suggest the transformer-based architectures may be well-suited for simple and texture-based tasks but less effective in scenarios requiring spatial reasoning across components. This limitation may stem from \texttt{CLIP}'s training data, which lacks industrial assembly images, and its default preprocessing, which can obscure peripheral or relational information critical to classification in complex assemblies.
\end{itemize}

\subsection{Advice to Practitioners}

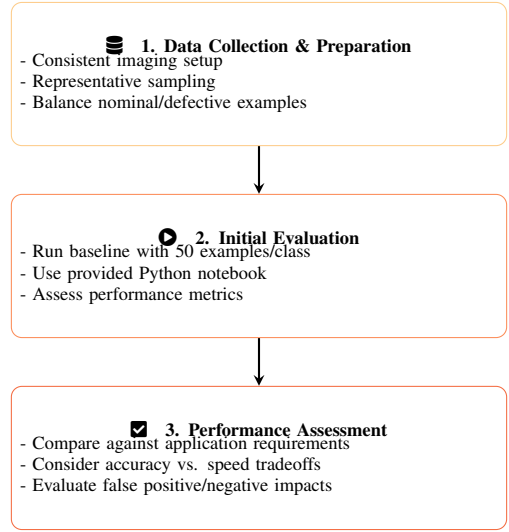
\begin{figure}[htb!]
\centering

\begin{tikzpicture}[
    box/.style={
        rectangle,
        draw,
        text width=2.5in,
        minimum height=.75in,
        align=center,
        font=\scriptsize, 
        fill=white,
        rounded corners
    },
    bullet/.style={
        text width=2.5in, 
        align=left,       
        font=\scriptsize  
    },
    arrow/.style={
        thick,
        ->,
        >=stealth
    }
]

\node[box, draw={rgb,255:red,253;green,204;blue,138}] (data) {\faDatabase\quad \textbf{1. Data Collection \& Preparation}\\

\vspace{-\baselineskip}

\begin{flushleft}
\scriptsize
- Consistent imaging setup\\
- Representative sampling\\
- Balance nominal/defective examples
\end{flushleft}};

\node[box, draw={rgb,255:red,252;green,141;blue,89}, below=0.25in of data] (initial) {\faPlayCircle\quad \textbf{2. Initial Evaluation}\\

\vspace{-\baselineskip}

\begin{flushleft}
\scriptsize
- Run baseline with 50 examples/class\\
- Use provided Python notebook\\
- Assess performance metrics
\end{flushleft}};

\node[box, draw={rgb,255:red,244;green,109;blue,67}, below=0.25in of initial] (decision) {\faCheckSquare\quad \textbf{3. Performance Assessment}\\

\vspace{-\baselineskip}

\begin{flushleft}
\scriptsize
- Compare against application requirements\\
- Consider accuracy vs. speed tradeoffs\\
- Evaluate false positive/negative impacts
\end{flushleft}};

\draw[arrow] (data) -- (initial);
\draw[arrow] (initial) -- (decision);

\end{tikzpicture}
\caption{Implementation workflow for CLIP-based quality control systems.}
\label{fig:implementation_workflow}
\end{figure}

Based on our findings and implementation experience, we propose a workflow for practitioners interested in deploying \texttt{CLIP}-based quality control systems (Figure \ref{fig:implementation_workflow}). The workflow begins with data collection and preparation, emphasizing consistent imaging conditions across samples. We recommend standardizing camera positions and lighting conditions while maintaining the highest possible image resolution for the application. For the learning set, collect at least 50 examples per class through stratified random sampling across different production runs, shifts, and expected operating conditions to capture natural process variation. When dealing with multiple defect types (as in our extrusion profile case), maintain equal proportions across defect categories. Additionally, prepare a similarly sized independent test set using the same sampling strategy to validate the model's performance.

The initial evaluation should leverage the provided Python notebook (available in the online materials) to assess the feasibility of using \texttt{CLIP}-based inspection for the intended application. Practitioners only need to update the file paths to their training and testing datasets to quickly determine whether \texttt{CLIP}-based few-shot learning meets their application requirements or aligns with the performance of existing approaches. If the initial results are promising but fall short of expectations, performance can be enhanced by increasing the number of few-shot examples (as illustrated in our extrusion and STS cases) or by transitioning to the \texttt{ViT-L/14} encoder model. This evaluation process requires minimal setup and provides clear guidance on whether to refine and research few-shot learning with \texttt{CLIP} or proceed to full implementation.

\subsection{Open Research Questions}

Below, we highlight three directions for future research:
\begin{enumerate}[label=(\arabic*), nosep]
    \item \textbf{Exploration of more advanced vision-language models:} Beyond our comparison of \texttt{ViT-L/14} and \texttt{ViT-B/32}, investigating other encoder models could unveil more effective architectures for manufacturing quality control applications. These comparisons should evaluate both performance metrics and computational efficiency to provide actionable insights for industrial implementations.

    \item \textbf{Enhancing performance in complex scenarios:} Addressing challenging cases, like the pipe staple example, represents another critical research direction. One promising strategy is combining visual and textual embeddings through frameworks like \texttt{Tip-Adapter} \citep{zhang2022tip}, enabling more nuanced feature extraction. Alternatively, integrating advanced segmentation models, such as the Segment Anything Model 2 \citep{ravi2024sam}, could isolate relevant components before applying the few-shot learning approach with \texttt{CLIP} for classification. While our current method's simplicity offers practical advantages, these hybrid approaches might be needed for tackling intricate manufacturing environments.

    \item \textbf{Monitoring \texttt{CLIP} probability outputs with control charts:} Employing simple univariate control charting techniques to monitor the nominal class probabilities derived from our \texttt{CLIP}-based approach over time presents a third research opportunity. This strategy could be particularly valuable in manufacturing settings where out-of-control behavior manifests as a gradual drift rather than an abrupt step change.

\end{enumerate}

\subsection {Limitations}
Our adaptation of the \texttt{CLIP} model demonstrates potential as a practical baseline for image inspection in manufacturing quality control.  However, we acknowledge several limitations that should guide the future use of this method.
\begin{itemize}
    \item{\textbf{Preprocessing Constraints:} \texttt{CLIP} requires square image inputs resized to fixed dimensions determined by the encoder model.  Although cropping images is common practice in image analysis, it may exclude certain peripheral defects or distort spatial context.}
    \item{\textbf{No Use of Text Embeddings:} Although \texttt{CLIP} supports image-text pairs in a shared embedding space, we did not leverage the text embeddings in our example. Our approach reflects real-world constraints in industrial vision systems where labeling images is costly.  Thus, our approach may underutilize \texttt{CLIP}'s full capabilities.}
    \item{\textbf{Generalizability:} Our five case studies span an array of quality control contexts; however, they do not represent the full diversity of industrial inspection tasks. By using the \texttt{CLIP} model as a baseline in future developments of application specific methodologies, we hope the understanding of the applicability of vision transformer models will expand.}
\end{itemize}

\section{Concluding Remarks}
\label{sec:conc}
Our expository study demonstrates that our \texttt{CLIP}-based few-shot learning approach can serve as an effective baseline for image-based quality control in manufacturing. Through five case studies, we showed that \texttt{CLIP} can achieve high classification accuracy with relatively small learning sets, often requiring only 50-100 examples per class compared to the thousands typically needed for deep learning approaches. The microstructure case study further highlighted the model's ability to adapt to multi-class classification tasks without retraining, demonstrating promising performance across multiple levels of classification granularity (from nominal vs. defective images to identifying specific defect types and their severities).

Our comprehensive comparison across all five case studies with MobileNetV2-based few-shot prototyping and AutoML Vision revealed that \texttt{CLIP} consistently delivered competitive performance while offering significant practical advantages in terms of implementation simplicity, runtime efficiency, and cost-effectiveness. Overall, the \texttt{CLIP}-based few-shot learning approach excelled at single-component inspection, texture analysis, and microstructure classification. However, its performance was poor in complex multi-component scenes, where more advanced computer vision and/or statistical monitoring approaches were needed. Our findings establish \texttt{CLIP}-based few-shot learning as a practical first baseline for quality engineers and researchers. It offers a simple yet powerful solution that should be evaluated before pursuing more complex implementations.

\section*{Data and Code Availability}
To facilitate reproducibility and encourage further research in this area, we provide comprehensive materials in our GitHub repository. The repository contains three main components: 
\begin{enumerate}[label=(\arabic*), nosep]
    \item a \ul{data} folder with separate subfolders for each case study, organized to clearly distinguish between training and testing datasets, with images categorized by their condition (nominal vs. defective) and defect type where applicable;
    \item a \ul{notebook} folder containing our Python implementation using the \texttt{CLIP} model, which we will share as a GitHub Gist (once the paper is accepted) for easy integration with platforms like Google Colab or DeepNote; and
    \item a \ul{results} folder storing all experimental outputs, including confusion matrices, animated visualizations of classification performance across different learning set sizes, and both raw and aggregated classification metrics in CSV format.
\end{enumerate}

\noindent Our repository includes the \faRProject\ \ code used to generate the stochastic textured surfaces dataset. These materials are hosted in the GitHub repository (\url{https://github.com/fmegahed/qe_genai}).

To upload the Python notebook from GitHub to Google Colab, start by opening Google Colab in a web browser. Once the Colab interface is open, click the \textit{File} menu located in the top left corner of the screen. From the dropdown options, select \textit{Upload notebook}, which will bring up a dialog box for choosing the source of the notebook. In this dialog, switch to the \textit{GitHub} tab and paste the URL of the GitHub repository: \url{https://github.com/fmegahed/}. After pasting the URL, press Enter or click the Search button to let Colab locate the notebooks in the repository. Once the repository list appears, locate and select \textit{fmegahed/qe\_genai}. Navigate to the "Path" section and find the notebook titled \textit{notebook/image\_inspection\_with\_clip.ipynb}. Click on the notebook file to load it into Colab. After the notebook is loaded, we can start working on it immediately. Colab allows us to edit, run, and interact with the notebook as needed.
 
\vspace{\baselineskip}
\noindent \textbf{Acknowledgement} \\
\noindent
The work of Hongyue Sun is partially supported by the NSF Grant No. FM-2134409 and CMMI-2412678. The research of Prof. Bianca Maria Colosimo and Prof. Marco Grasso was supported by ACCORDO Quadro ASI-POLIMI “Attività di Ricerca e Innovazione” n. 2018-5-HH.0, collaboration agreement between the Italian Space Agency and Politecnico di Milano.

\bibliographystyle{chicago}
\bibliography{refs}

\begin{thebibliography}{}

\bibitem[\protect\citeauthoryear{Barrett, Eghtesad, McCabe, Clausen, Brown,
  Vogel, and Knezevic}{Barrett et~al.}{2019}]{barrett2019generalized}
Barrett, T.~J., A.~Eghtesad, R.~J. McCabe, B.~Clausen, D.~W. Brown, S.~C.
  Vogel, and M.~Knezevic (2019).
\newblock A generalized spherical harmonics-based procedure for the
  interpolation of partial datasets of orientation distributions to enable
  crystal mechanics-based simulations.
\newblock {\em Materialia\/}~{\em 6}, 100328.

\bibitem[\protect\citeauthoryear{Bostanabad, Zhang, Li, Kearney, Brinson,
  Apley, Liu, and Chen}{Bostanabad et~al.}{2018}]{bostanabad2018computational}
Bostanabad, R., Y.~Zhang, X.~Li, T.~Kearney, L.~C. Brinson, D.~W. Apley, W.~K.
  Liu, and W.~Chen (2018).
\newblock Computational microstructure characterization and reconstruction:
  Review of the state-of-the-art techniques.
\newblock {\em Progress in Materials Science\/}~{\em 95}, 1--41.

\bibitem[\protect\citeauthoryear{Brandon and Kaplan}{Brandon and
  Kaplan}{2013}]{Brandon&Kaplan2013}
Brandon, D. and W.~Kaplan (2013).
\newblock {\em Microstructural characterization of Materials}.
\newblock John Wiley \& Sons.

\bibitem[\protect\citeauthoryear{Bridle}{Bridle}{1990}]{bridle1990probabilistic}
Bridle, J.~S. (1990).
\newblock Probabilistic interpretation of feedforward classification network
  outputs, with relationships to statistical pattern recognition.
\newblock In F.~F. Souli{\'e} and J.~H{\'e}rault (Eds.), {\em Neurocomputing},
  Berlin, Heidelberg, pp.\  227--236. Springer Berlin Heidelberg.

\bibitem[\protect\citeauthoryear{Bui and Apley}{Bui and
  Apley}{2018}]{bui2018monitoring}
Bui, A.~T. and D.~W. Apley (2018).
\newblock A monitoring and diagnostic approach for stochastic textured
  surfaces.
\newblock {\em Technometrics\/}~{\em 60\/}(1), 1--13.

\bibitem[\protect\citeauthoryear{Bui and Apley}{Bui and
  Apley}{2021}]{bui2021spc4sts}
Bui, A.~T. and D.~W. Apley (2021).
\newblock spc4sts: Statistical process control for stochastic textured surfaces
  in {R}.
\newblock {\em Journal of Quality Technology\/}~{\em 53\/}(3), 219--242.

\bibitem[\protect\citeauthoryear{Carvalho, Lafou, Durupt, Leblanc, and
  Grandvalet}{Carvalho et~al.}{2024}]{carvalho2024detecting}
Carvalho, P., M.~Lafou, A.~Durupt, A.~Leblanc, and Y.~Grandvalet (2024).
\newblock Detecting visual anomalies in an industrial environment: Unsupervised
  methods put to the test on the {AutoVI} dataset.
\newblock {\em Computers in Industry\/}~{\em 163}, 104151.

\bibitem[\protect\citeauthoryear{Colosimo}{Colosimo}{2018}]{colosimo2018modeling}
Colosimo, B.~M. (2018).
\newblock Modeling and monitoring methods for spatial and image data.
\newblock {\em Quality Engineering\/}~{\em 30\/}(1), 94--111.

\bibitem[\protect\citeauthoryear{Colosimo, Huang, Dasgupta, and Tsung}{Colosimo
  et~al.}{2018}]{colosimo2018opportunities}
Colosimo, B.~M., Q.~Huang, T.~Dasgupta, and F.~Tsung (2018).
\newblock Opportunities and challenges of quality engineering for additive
  manufacturing.
\newblock {\em Journal of Quality Technology\/}~{\em 50\/}(3), 233--252.

\bibitem[\protect\citeauthoryear{Deng, Dong, Socher, Li, Li, and Fei-Fei}{Deng
  et~al.}{2009}]{deng2009imagenet}
Deng, J., W.~Dong, R.~Socher, L.-J. Li, K.~Li, and L.~Fei-Fei (2009).
\newblock Imagenet: A large-scale hierarchical image database.
\newblock In {\em 2009 IEEE conference on computer vision and pattern
  recognition}, pp.\  248--255. Ieee.

\bibitem[\protect\citeauthoryear{Gola, Britz, Staudt, Winter, Schneider,
  Ludovici, and M{\"u}cklich}{Gola et~al.}{2018}]{gola2018advanced}
Gola, J., D.~Britz, T.~Staudt, M.~Winter, A.~S. Schneider, M.~Ludovici, and
  F.~M{\"u}cklich (2018).
\newblock Advanced microstructure classification by data mining methods.
\newblock {\em Computational Materials Science\/}~{\em 148}, 324--335.

\bibitem[\protect\citeauthoryear{{Google Cloud}}{{Google
  Cloud}}{2025a}]{vertexai-nas-overview2025}
{Google Cloud} (2025a, June).
\newblock About {Vertex AI} neural architecture search.
\newblock
  \url{https://cloud.google.com/vertex-ai/docs/training/neural-architecture-search/overview}.
\newblock Accessed: 2025-06-15.

\bibitem[\protect\citeauthoryear{{Google Cloud}}{{Google
  Cloud}}{2025b}]{vertexai-beginnersguide2025}
{Google Cloud} (2025b, June).
\newblock Automl beginner's guide.
\newblock
  \url{https://cloud.google.com/vertex-ai/docs/beginner/beginners-guide}.
\newblock Accessed: 2025-06-15.

\bibitem[\protect\citeauthoryear{{Google Cloud}}{{Google
  Cloud}}{2025c}]{vertexai-searchspaces2025}
{Google Cloud} (2025c, June).
\newblock Prebuilt search spaces.
\newblock
  \url{https://cloud.google.com/vertex-ai/docs/training/neural-architecture-search/search-spaces}.
\newblock Accessed: 2025-06-15.

\bibitem[\protect\citeauthoryear{Grandini, Bagli, and Visani}{Grandini
  et~al.}{2020}]{grandini2020metrics}
Grandini, M., E.~Bagli, and G.~Visani (2020).
\newblock Metrics for multi-class classification: an overview.
\newblock {\em arXiv preprint arXiv:2008.05756\/}.

\bibitem[\protect\citeauthoryear{He, Zhang, Ren, and Sun}{He
  et~al.}{2016}]{he2016deep}
He, K., X.~Zhang, S.~Ren, and J.~Sun (2016).
\newblock Deep residual learning for image recognition.
\newblock In {\em Proceedings of the IEEE Conference on Computer Vision and
  Pattern Recognition}, pp.\  770--778.

\bibitem[\protect\citeauthoryear{He, Zuo, Zhang, and Megahed}{He
  et~al.}{2016}]{he2016image}
He, Z., L.~Zuo, M.~Zhang, and F.~M. Megahed (2016).
\newblock An image-based multivariate generalized likelihood ratio control
  chart for detecting and diagnosing multiple faults in manufactured products.
\newblock {\em International Journal of Production Research\/}~{\em 54\/}(6),
  1771--1784.

\bibitem[\protect\citeauthoryear{Kang, Jiao, Geng, and Nagarajan}{Kang
  et~al.}{2024}]{kang2024modern}
Kang, Y., Y.~Jiao, X.~Geng, and M.~Nagarajan (2024).
\newblock Deep vision in smart manufacturing: {MODERN} framework for
  intelligent quality monitoring and diagnosis.
\newblock Research Paper 4856345, University of Miami Business School.
  Available at \url{https://dx.doi.org/10.2139/ssrn.4856345}.

\bibitem[\protect\citeauthoryear{Krizhevsky, Sutskever, and Hinton}{Krizhevsky
  et~al.}{2012}]{krizhevsky2012imagenet}
Krizhevsky, A., I.~Sutskever, and G.~E. Hinton (2012).
\newblock Imagenet classification with deep convolutional neural networks.
\newblock In F.~Pereira, C.~Burges, L.~Bottou, and K.~Weinberger (Eds.), {\em
  Advances in Neural Information Processing Systems}, Volume~25. Curran
  Associates, Inc.

\bibitem[\protect\citeauthoryear{Larmuseau, Sluydts, Theuwissen, Duprez,
  Dhaene, and Cottenier}{Larmuseau et~al.}{2021}]{larmuseau2021race}
Larmuseau, M., M.~Sluydts, K.~Theuwissen, L.~Duprez, T.~Dhaene, and
  S.~Cottenier (2021).
\newblock Race against the machine: can deep learning recognize microstructures
  as well as the trained human eye?
\newblock {\em Scripta Materialia\/}~{\em 193}, 33--37.

\bibitem[\protect\citeauthoryear{Lever, Krzywinski, and Altman}{Lever
  et~al.}{2016}]{lever2016classification}
Lever, J., M.~Krzywinski, and N.~Altman (2016).
\newblock Classification evaluation.
\newblock {\em Nature Methods\/}~{\em 13\/}(8), 603--604.

\bibitem[\protect\citeauthoryear{Liu, Zhao, Wang, Yan, Yang, Chen, Lu, and
  Lu}{Liu et~al.}{2022}]{liu2022additive}
Liu, Z., D.~Zhao, P.~Wang, M.~Yan, C.~Yang, Z.~Chen, J.~Lu, and Z.~Lu (2022).
\newblock Additive manufacturing of metals: Microstructure evolution and
  multistage control.
\newblock {\em Journal of Materials Science \& Technology\/}~{\em 100},
  224--236.

\bibitem[\protect\citeauthoryear{Megahed and Camelio}{Megahed and
  Camelio}{2012}]{megahed2012real}
Megahed, F.~M. and J.~A. Camelio (2012).
\newblock Real-time fault detection in manufacturing environments using face
  recognition techniques.
\newblock {\em Journal of Intelligent Manufacturing\/}~{\em 23}, 393--408.

\bibitem[\protect\citeauthoryear{Megahed, Chen, Jones-Farmer, Rigdon,
  Krzywinski, and Altman}{Megahed et~al.}{2024}]{megahed2024comparing}
Megahed, F.~M., Y.-J. Chen, L.~A. Jones-Farmer, S.~E. Rigdon, M.~Krzywinski,
  and N.~Altman (2024).
\newblock Comparing classifier performance with baselines.
\newblock {\em Nature Methods\/}~{\em 21\/}(4), 546–--548.

\bibitem[\protect\citeauthoryear{Megahed, Chen, Megahed, Ong, Altman, and
  Krzywinski}{Megahed et~al.}{2021}]{megahed2021class}
Megahed, F.~M., Y.-J. Chen, A.~Megahed, Y.~Ong, N.~Altman, and M.~Krzywinski
  (2021).
\newblock The class imbalance problem.
\newblock {\em Nature Methods\/}~{\em 18\/}(11), 1270--1272.

\bibitem[\protect\citeauthoryear{Megahed, Wells, Camelio, and Woodall}{Megahed
  et~al.}{2012}]{megahed2012spatiotemporal}
Megahed, F.~M., L.~J. Wells, J.~A. Camelio, and W.~H. Woodall (2012).
\newblock A spatiotemporal method for the monitoring of image data.
\newblock {\em Quality and Reliability Engineering International\/}~{\em
  28\/}(8), 967--980.

\bibitem[\protect\citeauthoryear{Megahed, Woodall, and Camelio}{Megahed
  et~al.}{2011}]{megahed2011review}
Megahed, F.~M., W.~H. Woodall, and J.~A. Camelio (2011).
\newblock A review and perspective on control charting with image data.
\newblock {\em Journal of Quality Technology\/}~{\em 43\/}(2), 83--98.

\bibitem[\protect\citeauthoryear{Menafoglio, Grasso, Secchi, and
  Colosimo}{Menafoglio et~al.}{2018}]{menafoglio2018profile}
Menafoglio, A., M.~Grasso, P.~Secchi, and B.~M. Colosimo (2018).
\newblock Profile monitoring of probability density functions via simplicial
  functional {PCA} with application to image data.
\newblock {\em Technometrics\/}~{\em 60\/}(4), 497--510.

\bibitem[\protect\citeauthoryear{Okhrin, Schmid, and Semeniuk}{Okhrin
  et~al.}{2024}]{okhrin2024control}
Okhrin, Y., W.~Schmid, and I.~Semeniuk (2024).
\newblock A control chart for monitoring image processes based on convolutional
  neural networks.
\newblock {\em Statistica Neerlandica\/}~{\em Early View}, 1--26.

\bibitem[\protect\citeauthoryear{OpenAI}{OpenAI}{2022}]{openai2021clipmodelcard}
OpenAI (2022).
\newblock {CLIP} model card.
\newblock Available at
  \url{https://github.com/openai/CLIP/blob/main/model-card.md}.
\newblock Accessed: 2025-01-06.

\bibitem[\protect\citeauthoryear{Pratap and Sardana}{Pratap and
  Sardana}{2022}]{pratap2022machine}
Pratap, A. and N.~Sardana (2022).
\newblock Machine learning-based image processing in materials science and
  engineering: A review.
\newblock {\em Materials Today: Proceedings\/}~{\em 62}, 7341--7347.

\bibitem[\protect\citeauthoryear{Radford, Kim, Hallacy, Ramesh, Goh, Agarwal,
  Sastry, Askell, Mishkin, Clark, et~al.}{Radford
  et~al.}{2021}]{radford2021learning}
Radford, A., J.~W. Kim, C.~Hallacy, A.~Ramesh, G.~Goh, S.~Agarwal, G.~Sastry,
  A.~Askell, P.~Mishkin, J.~Clark, et~al. (2021).
\newblock Learning transferable visual models from natural language
  supervision.
\newblock In {\em International Conference on Machine Learning}, pp.\
  8748--8763. PMLR.

\bibitem[\protect\citeauthoryear{Ravi, Gabeur, Hu, Hu, Ryali, Ma, Khedr,
  R{\"a}dle, Rolland, Gustafson, et~al.}{Ravi et~al.}{2024}]{ravi2024sam}
Ravi, N., V.~Gabeur, Y.-T. Hu, R.~Hu, C.~Ryali, T.~Ma, H.~Khedr, R.~R{\"a}dle,
  C.~Rolland, L.~Gustafson, et~al. (2024).
\newblock {SAM 2}: Segment anything in images and videos.
\newblock {\em arXiv preprint arXiv:2408.00714v2\/}.

\bibitem[\protect\citeauthoryear{Sandler, Howard, Zhu, Zhmoginov, and
  Chen}{Sandler et~al.}{2018}]{sandler2018mobilenetv2}
Sandler, M., A.~Howard, M.~Zhu, A.~Zhmoginov, and L.-C. Chen (2018).
\newblock Mobilenetv2: Inverted residuals and linear bottlenecks.
\newblock In {\em Proceedings of the IEEE conference on computer vision and
  pattern recognition}, pp.\  4510--4520.

\bibitem[\protect\citeauthoryear{Simonyan and Zisserman}{Simonyan and
  Zisserman}{2015}]{simonyan2015very}
Simonyan, K. and A.~Zisserman (2015).
\newblock Very deep convolutional networks for large-scale image recognition.
\newblock In {\em 3rd International Conference on Learning Representations
  (ICLR 2015)}, pp.\  1--14. Computational and Biological Learning Society.

\bibitem[\protect\citeauthoryear{Tan and Le}{Tan and
  Le}{2019}]{tan2019efficientnet}
Tan, M. and Q.~Le (2019).
\newblock Efficientnet: Rethinking model scaling for convolutional neural
  networks.
\newblock In {\em International Conference on Machine Learning}, pp.\
  6105--6114. PMLR.

\bibitem[\protect\citeauthoryear{Wei, Grasso, Bisheh, Paynabar, and
  Colosimo}{Wei et~al.}{2025}]{wei2025lowdimensional}
Wei, Y., M.~Grasso, M.~N. Bisheh, K.~Paynabar, and B.~M. Colosimo (2025).
\newblock A novel low-dimensional learning approach for automated
  classification of microstructure data with application to additive
  manufacturing.
\newblock Under review.

\bibitem[\protect\citeauthoryear{Yan, Paynabar, and Shi}{Yan
  et~al.}{2018}]{yan2018real}
Yan, H., K.~Paynabar, and J.~Shi (2018).
\newblock Real-time monitoring of high-dimensional functional data streams via
  spatio-temporal smooth sparse decomposition.
\newblock {\em Technometrics\/}~{\em 60\/}(2), 181--197.

\bibitem[\protect\citeauthoryear{Yang, Grasso, Bisheh, Paynabar, and M.}{Yang
  et~al.}{2025}]{WeietAl2025}
Yang, W., M.~Grasso, M.~N. Bisheh, K.~Paynabar, and C.~B. M. (2025).
\newblock Automated classification of microstructure data in additive
  manufacturing via low-dimensional learning.
\newblock Manuscript under review.

\bibitem[\protect\citeauthoryear{Zhang, Hu, Li, Huang, Deng, Qiao, Gao, and
  Li}{Zhang et~al.}{2023}]{zhang2023prompt}
Zhang, R., X.~Hu, B.~Li, S.~Huang, H.~Deng, Y.~Qiao, P.~Gao, and H.~Li (2023).
\newblock Prompt, generate, then cache: Cascade of foundation models makes
  strong few-shot learners.
\newblock In {\em Proceedings of the IEEE/CVF Conference on Computer Vision and
  Pattern Recognition}, pp.\  15211--15222.

\bibitem[\protect\citeauthoryear{Zhang, Zhang, Fang, Gao, Li, Dai, Qiao, and
  Li}{Zhang et~al.}{2022}]{zhang2022tip}
Zhang, R., W.~Zhang, R.~Fang, P.~Gao, K.~Li, J.~Dai, Y.~Qiao, and H.~Li (2022).
\newblock Tip-adapter: Training-free adaption of clip for few-shot
  classification.
\newblock In {\em European Conference on Computer Vision}, pp.\  493--510.
  Springer.

\end{thebibliography}

\end{document}